\documentclass[runningheads]{llncs}

\usepackage{comment}
\usepackage{amsmath,amssymb} 
\usepackage[accsupp]{axessibility}  
\usepackage[utf8]{inputenc} 
\usepackage[T1]{fontenc}    
\usepackage{url}            
\usepackage{booktabs}       
\usepackage[font=small,labelfont=bf]{caption}
\usepackage{subcaption}
\usepackage{amsfonts}       
\usepackage{nicefrac}       
\usepackage{microtype}      
\usepackage{times}
\usepackage{epsfig}
\usepackage[T1]{fontenc}
\usepackage{graphicx}
\usepackage{tabu}
\usepackage{textcomp}
\usepackage{gensymb}
\usepackage{wrapfig}
\usepackage{xcolor}
\usepackage{enumitem}
\usepackage[splitrule]{footmisc}
\usepackage{siunitx}
\usepackage{bm}
\usepackage{dirtytalk}
\usepackage{mathtools}
\usepackage{multirow}
\usepackage{wrapfig}
\usepackage{float}

\usepackage[square,comma,numbers,sort&compress]{natbib}

\makeatletter 
\renewcommand\@biblabel[1]{#1.} 
\makeatother

\usepackage[pagebackref,breaklinks,colorlinks]{hyperref}
\hypersetup{
    linkcolor=cyan,
    urlcolor=cyan,
}
\usepackage[capitalize]{cleveref}
\crefname{section}{Sec.}{Secs.}
\Crefname{section}{Section}{Sections}
\Crefname{table}{Table}{Tables}
\crefname{table}{Tab.}{Tabs.}

\usepackage{orcidlink}

\let\emptyset\varnothing
\newcommand{\tobs}{\ensuremath{\bm{t}_\mathrm{obs}}}
\newcommand{\tfut}{\ensuremath{\bm{t}_\mathrm{fut}}}
\newcommand{\norm}[1]{\left\lVert #1 \right\rVert}
\newcommand{\ra}[1]{\renewcommand{\arraystretch}{#1}}

\begin{document}
\pagestyle{headings}
\mainmatter
\def\ECCVSubNumber{317}  

\title{Social Processes: Self-Supervised Meta-Learning\\over Conversational Groups\\ for Forecasting Nonverbal Social Cues}

\titlerunning{Social Processes}
%
\author{Chirag Raman\inst{1}\orcidlink{0000-0003-4894-4206} \and
Hayley Hung\inst{1}\orcidlink{0000-0001-9574-5395} \and
Marco Loog\inst{1,2}\orcidlink{0000-0002-1298-8461}}
\authorrunning{C. Raman et al.}
%
\institute{Delft University of Technology, Delft, The Netherlands \and
University of Copenhagen, Copenhagen, Denmark\\
\email{\{c.a.raman, h.hung, m.loog\}@tudelft.nl}}
\maketitle

\begin{abstract}
Free-standing social conversations constitute a yet underexplored setting for human behavior forecasting. While the task of predicting pedestrian trajectories has received much recent attention, an intrinsic difference between these settings is how groups form and disband. Evidence from social psychology suggests that group members in a conversation explicitly self-organize to sustain the interaction by adapting to one another's behaviors. Crucially, the same individual is unlikely to adapt similarly across different groups; contextual factors such as perceived relationships, attraction, rapport, etc., influence the entire spectrum of participants' behaviors. A question arises: how can we jointly forecast the mutually dependent futures of conversation partners by modeling the dynamics unique to every group? In this paper, we propose the \textit{Social Process} (SP) models, taking a novel meta-learning and stochastic perspective of group dynamics. Training group-specific forecasting models hinders generalization to unseen groups and is challenging given limited conversation data. In contrast, our SP models treat interaction sequences from a single group as a meta-dataset: we condition forecasts for a sequence from a given group on other observed-future sequence pairs from the same group. In this way, an SP model learns to adapt its forecasts to the unique dynamics of the interacting partners, generalizing to unseen groups in a data-efficient manner. Additionally, we first rethink the task formulation itself, motivating task requirements from social science literature that prior formulations have overlooked. For our formulation of \textit{Social Cue Forecasting}, we evaluate the empirical performance of our SP models against both non-meta-learning and meta-learning approaches with similar assumptions. The SP models yield improved performance on synthetic and real-world behavior datasets.
\end{abstract}


\section{Introduction}
\label{sec:introduction}

Picture a conversing group of people in a free-standing social setting. To conduct such exchanges, we transfer high-order social signals across space and time through explicit low-level behavior cues\textemdash examples include our pose, gestures, gaze, and floor control actions \cite{kendonConductingInteractionPatterns1990, vinciarelli_social_2009, bohusModelsMultipartyEngagement2009}. Evidence suggests that we employ anticipation of these and other cues to navigate daily social interactions \cite{kendonConductingInteractionPatterns1990, ishiiPredictionNextUtteranceTiming2017, keitel2015use, garrod2015use, rochet2014take, Wlodarczak2016RespiratoryTC}.
Consequently, for machines to truly develop adaptive social skills, they need to have the ability to forecast the future. For instance, foreseeing the upcoming behaviors of partners in advance can enable interactive agents to choose more fluid interaction policies \cite{bohusManagingHumanRobotEngagement2014}, or contend with uncertainties in imperfect real-time inferences surrounding cues \cite{bohusModelsMultipartyEngagement2009}.

In literature, behavior forecasting works mainly consider data at two representations with an increasing level of abstraction: low-level cues or features that are extracted manually or automatically from raw audiovisual data, and manually labeled high-order events or actions. The forecasting task has primarily been formulated to predict future event or action labels from observed cues or other high-order event or action labels \cite{garrod2015use, keitel2015use, bohusManagingHumanRobotEngagement2014, vandoornRitualsLeavingPredictive2018, airaleSocialInteractionGANMultipersonInteraction2021, sanghvi2019mgpi, bilakhiaAudiovisualDetectionBehavioural2013}. Moreover, identifying patterns predictive of certain semantic events has been a long-standing topic of focus in the social sciences, where researchers primarily employ a top-down workflow. First, the events of interest are selected for consideration. Then their relationship to preceding cues or other high-order actions are studied in isolation through exploratory or confirmatory analysis \cite{liemPsychologyMeetsMachine2018, nilsenExploratoryConfirmatoryResearch2020}.
Examples of such semantic events include speaker turn transitions \cite{garrod2015use, keitel2015use}, mimicry episodes \cite{bilakhiaAudiovisualDetectionBehavioural2013}, the termination of an interaction \cite{bohusManagingHumanRobotEngagement2014, vandoornRitualsLeavingPredictive2018}, or high-order social actions \cite{airaleSocialInteractionGANMultipersonInteraction2021, sanghvi2019mgpi}.
\begin{figure*}[t!]
\makebox[\textwidth][c]{\includegraphics[width=\textwidth]{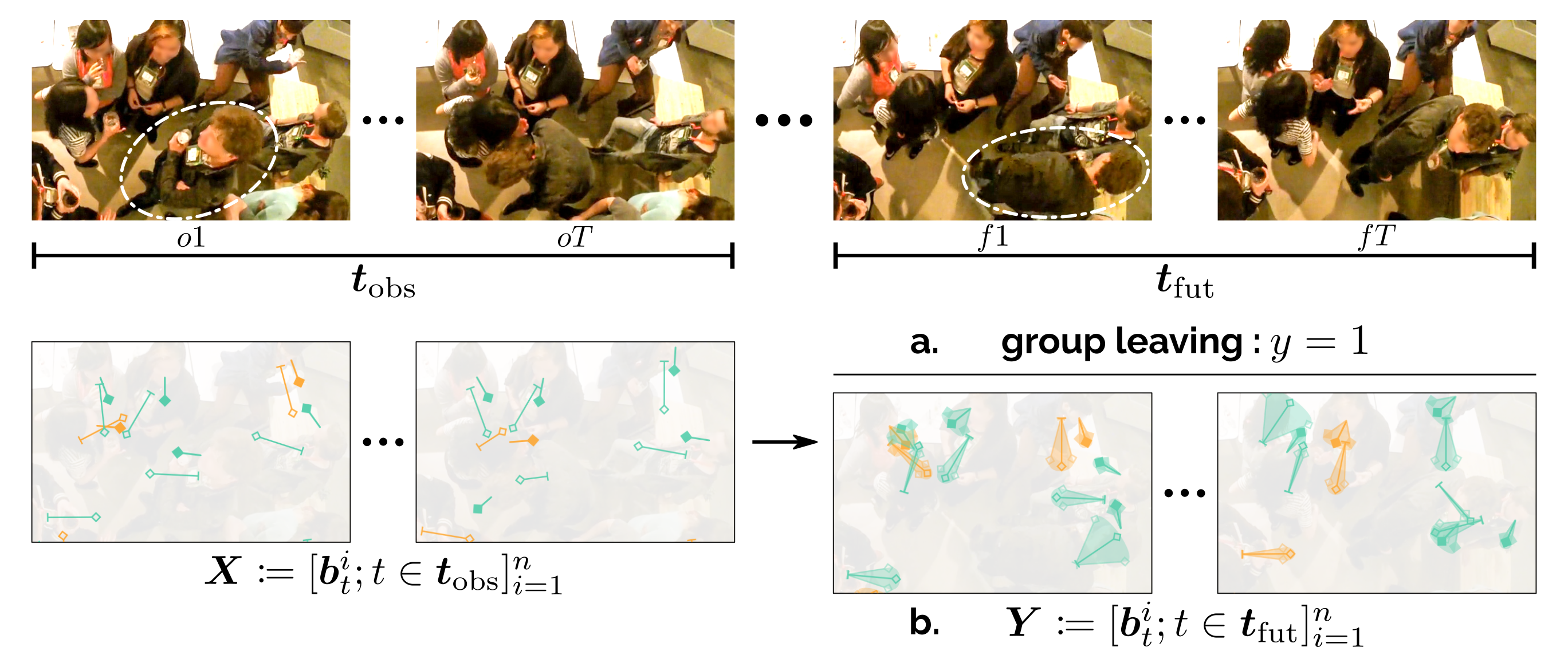}}
\caption{Conceptual illustration of forecasting approaches on an in-the-wild conversation from the MatchNMingle dataset \cite{cabrera2018matchnmingle}. \textbf{Top.} A \textit{group leaving} event \cite{vandoornRitualsLeavingPredictive2018}: the circled individual has moved from one  group in the observed window $\tobs \coloneqq [o1 \ldots oT]$ to another in a future window $\tfut \coloneqq [f1 \ldots fT]$.  \textbf{Bottom.} Input behavioral cues $\bm{b}^i_t$: head pose (solid normal), body pose (hollow normal), and speaking status (speaker in orange). \textbf{a.} The top-down approach entails predicting the event label from such cues over $\tobs$, from only $200$ instances of group leaving in over $90$ minutes of interaction \cite{vandoornRitualsLeavingPredictive2018}. \textbf{b.} Our proposed bottom-up, self-supervised formulation of \textit{Social Cue Forecasting} involves regressing a future distribution for the same low-level input cues over $\tfut$ (shaded spread). This enables utilizing the full $90$ minutes of event-unlabeled data. 
}
\label{fig:concept}
\end{figure*}

One hurdle in such a top-down paradigm is data efficiency.
The labeled events often occur infrequently over the interaction, reducing the effective amount of labeled data. This, combined with the fact that collecting behavior data is cost and labor-intensive, precludes the effective application of neural supervised learning techniques that tend to be data demanding. More recently, some approaches have adopted a more bottom-up formulation for dyadic conversations. The task entails predicting event-independent future cues for a single target participant or virtual avatar from the preceding observed cues of both participants \cite{palmero2022chalearn, ahujaReactNotReact2019}. Since training sequences are not limited to windows around semantic events, such a formulation is more data-efficient. \autoref{fig:concept} illustrates the top-down and bottom-up approaches conceptually.

In practice, however, the concrete formulations within the bottom-up paradigm \cite{palmero2022chalearn, ahujaReactNotReact2019} suffer from several conceptual problems: (i) predictions are made for a single individual using cues from both individuals as input; since people behave differently, this entails training one forecasting model per person; (ii) even so, predicting a future for one individual at a time is undesirable as these futures are not independent; and (iii) the prediction is only a single future, despite evidence that the future is not deterministic, and the same observed sequence can result in multiple socially-valid continuations \cite{heldnerPausesGapsOverlaps2010, duncanSignalsRulesTaking1972, mooreNonverbalCourtshipPatterns1985}. 

To address all these issues, we introduce a self-supervised forecasting task called Social Cue Forecasting: predicting a \textit{distribution} over future multimodal cues \textit{jointly for all group members} from their same preceding multimodal cues. 
Note that we use \textit{self-supervised} here to simply distinguish from the formulations where the predicted quantity (e.g. event-labels) is of a different representation than the observed input (e.g. cues). Given the cue data, the inputs and outputs of our formulation are both cues, so we \textit{obtain the supervisory signal from the data itself}.

Furthermore, a crucial characteristic of free-standing conversations is that people sustain the interaction by explicitly adapting to one another's behaviors \cite{kendonConductingInteractionPatterns1990}. Moreover, the way a person adapts to their partners is a function of several complex factors surrounding their interpersonal relationships and the social setting \citetext{\citealp[Chap.~1]{moore2013stacks}; \citealp[p.~237]{kendonConductingInteractionPatterns1990}}. The social dynamics guiding such behavior are embedded in the constellation of participant cues and are distinct for every unique grouping of individuals. As such, a model should adapt its forecasts to the group under consideration. (Even in the pedestrian setting where coordination is only implicit, \citet[Sec.~8.4.1]{rudenko2020human} observe that failing to adapt predictions to different individuals is still a limitation). For our methodological contribution, we propose the probabilistic Social Processes models, viewing each conversation group as a meta-learning \textit{task}. This allows for capturing social dynamics unique to each group without learning group-specific models and generalizing to unseen groups at evaluation in a data-efficient manner. We believe that this framing of SCF as a \textit{few-shot} function estimation problem is especially suitable for conversation forecasting\textemdash a limited data regime where good uncertainty estimates are desirable. Concretely, we make the following contributions:
\begin{itemize}[itemsep=0em]
    \item We introduce and formalize the novel task of Social Cue Forecasting (SCF), addressing the conceptual drawbacks of past formulations. 
    \item For SCF, we propose and evaluate the family of socially aware probabilistic Seq2Seq models we call Social Processes (SP). 
\end{itemize}

\section{Related Work} \label{sec:relatedwork}
To aid readers from different disciplines situate our work within the broader research landscape, we categorize behavior-forecasting literature by interaction focus \cite{goffmanBehaviorPublicPlaces1966}.  In a focused interaction, such as conversations, participants explicitly coordinate their behaviors to sustain the interaction. In unfocused interactions, coordination is implicit, such as when pedestrians avoid collisions.

\paragraph{Focused Interactions.} The predominant interest in conversation forecasting stems from the social sciences, with a focus on identifying patterns that are predictive of
upcoming speaking turns \cite{garrod2015use, keitel2015use, rochet2014take, Wlodarczak2016RespiratoryTC},
disengagement from an interaction \cite{bohusManagingHumanRobotEngagement2014, vandoornRitualsLeavingPredictive2018}, or the splitting or merging of groups \cite{wangGroupSplitMerge2020}.
Other works forecast the time-evolving size of a group \cite{mastrangeliRoundtableAbstractModel2010} or semantic social action labels \cite{airaleSocialInteractionGANMultipersonInteraction2021, sanghvi2019mgpi}. 
More recently, there has also been a growing interest in the computer vision community for tasks related to inferring low-level cues of participants either from their partners' cues \cite{jooSocialArtificialIntelligence2019} or raw multimodal sensor data \cite{tanMultimodalJointHead2021}. 
Here there has also been some interest in forecasting nonverbal behavior, mainly for dyadic interactions \cite{palmero2022chalearn, tuyen2022context, ahujaReactNotReact2019}. The task involves forecasting the future cues of a target individual from the preceding cues of both participants.

\paragraph{Unfocused Interactions.} 
Early approaches for forecasting pedestrian or vehicle trajectories were heuristic-based, involving hand-crafted energy potentials to describe the influence pedestrians and vehicles have on each other
\cite{helbingSocialForceModel1995, wasSocialDistancesModel2006, antoniniDiscreteChoiceModels2006, Treuille:2006:CC, robicquetLearningSocialEtiquette2016a, wangGaussianProcessDynamical2008, tayModellingSmoothPaths2007, pattersonIntentAwareProbabilisticTrajectory2019}.
Recent approaches build upon the idea of encoding relative positional information directly into a neural architecture \cite{alahiSocialLSTMHuman2016, zhangSRLSTMStateRefinement2019, guptaSocialGANSocially2018,hasanForecastingPeopleTrajectories2019,huangSTGATModelingSpatialTemporal2019, mohamedSocialSTGCNNSocialSpatioTemporal2020,zhaoTNTTargetdriveNTrajectory2020,gillesTHOMASTrajectoryHeatmap2022}. Some works go beyond locations, predicting keypoints in group activities \cite{yaoMultipleGranularityGroup2018, adeli2020socially}. \citet{rudenko2020human} provide a survey of approaches within this space. 

\paragraph{Non-Interaction Settings.} 
Here, the focus has been on forecasting individual poses from images \cite{chaoForecastingHumanDynamics2017} and video \cite{fragkiadakiRecurrentNetworkModels2015, walkerPoseKnowsVideo2017}, or synthesizing poses using high-level control parameters \cite{habibieRecurrentVariationalAutoencoder2017, pavlloQuaterNetQuaternionbasedRecurrent2018}. The self-supervised aspects of our task formulation are related to visual forecasting, where the goal has been to predict non-semantic low-level pixel features or intermediate representations \cite{ranzatoVideoLanguageModeling2014, walkerPoseKnowsVideo2017, walkerDenseOpticalFlow2015, dosovitskiyFlowNetLearningOptical2015, robicquetLearningSocialEtiquette2016a, walkerPatchFutureUnsupervised2014, vondrickAnticipatingVisualRepresentations2016}. Such learned representations have been utilized for other tasks like semi-supervised classification \cite{srivastavaUnsupervisedLearningVideo2015}, or training agents in immersive environments \cite{dosovitskiyLearningActPredicting2016}.

\medskip
\noindent For the interested reader, we further discuss practical considerations distinguishing forecasting in conversation and pedestrian settings in Appendix~\ref{app:review}. 

\section{Social Cue Forecasting: Task Formalization} \label{sec:task}

While self-supervision has shown promise for learning representations of language and video data, is this bottom-up approach conceptually reasonable for behavior cues? The crucial observation we make is that the semantic meaning transferred in interactions (the so-called \textit{social signal} \cite{ambady2000toward}) is already embedded in the 
low-level cues \cite{vinciarelliSocialSignalProcessing2009a}. So representations of this high-level semantic meaning that we associate with actions and events (e.g. \textit{group leaving}) can be learned from the low-level dynamics in the cues. 

\subsection{Formalization and Distinction from Prior Task Formulations} \label{sec:task-reqs}
The objective of SCF is to predict future behavioral cues of \textit{all} people involved in a social encounter given an observed sequence of their behavioral features. Formally, let us denote a window of monotonically increasing observed timesteps as $\tobs~\coloneqq~[o1, o2, ..., oT]$, and an unobserved future time window as $\tfut~\coloneqq~[f1, f2, ..., fT]$, $f1>oT$. Note that
$\tfut$ and $\tobs$ can be of different lengths, and $\tfut$ need not immediately follow $\tobs$.
Given $n$ interacting participants, let us denote their social cues over $\tobs$ and $\tfut$ as
\begin{subequations}
\begin{gather}
    \bm{X} \coloneqq [\bm{b}^i_{t}; t \in \tobs]_{i=1}^n,\quad
    \bm{Y} \coloneqq [\bm{b}^i_{t}; t \in \tfut]_{i=1}^n. \tag{\theequation a, b}
\end{gather}
\end{subequations}
The vector $\bm{b}^i_{t}$ encapsulates the multimodal cues of interest from participant $i$ at time $t$. These can include head and body pose, speaking status, facial expressions, gestures, verbal content\textemdash any information streams that combine to transfer social meaning.

\paragraph{Distribution over Futures.}
In its simplest form, given an $\bm{X}$, the objective of SCF is to learn a single function $f$ such that $\bm{Y} = f(\bm{X})$. However, an inherent challenge in forecasting behavior is that
an observed sequence of interaction does not have a deterministic future and can result in multiple socially valid ones\textemdash
a window of overlapping speech between people may and may not
result in a change of speaker \cite{heldnerPausesGapsOverlaps2010, duncanSignalsRulesTaking1972}, a change in head orientation may continue into a sweeping
glance across the room or a darting glance stopping at a
recipient of interest \cite{mooreNonverbalCourtshipPatterns1985}.
In some cases, certain observed behaviors\textemdash intonation and gaze cues
\cite{keitel2015use, kalmaGazingTriadsPowerful1992} or synchronization in speaker-listener speech
\cite{levinsonTimingTurntakingIts2015} for turn-taking\textemdash
may make some outcomes more likely than others. Given that there are both
supporting and challenging arguments for how these observations influence subsequent behaviors
\citetext{\citealp[p.~5]{levinsonTimingTurntakingIts2015}; \citealp[p.~22]{kalmaGazingTriadsPowerful1992}},
it would be beneficial if a data-driven model expresses a measure of uncertainty in its forecasts.
We do this by modeling the distribution over possible futures $p(\bm{Y}|\bm{X})$, rather than a single future $\bm{Y}$ for a given $\bm{X}$, the latter being the case for previous formulations for cues \cite{jooSocialArtificialIntelligence2019, ahujaReactNotReact2019, yaoMultipleGranularityGroup2018} and actions \cite{sanghvi2019mgpi, airaleSocialInteractionGANMultipersonInteraction2021}.

\paragraph{Joint Modeling of Future Uncertainty.}
A defining characteristic of focused interactions is that the 
participants sustain the shared interaction through explicit, cooperative coordination of behavior \citetext{\citealp[p.~220]{kendonConductingInteractionPatterns1990}}\textemdash the futures of interacting individuals are not independent given an observed window of group behavior. It is therefore essential to capture uncertainty in forecasts at the \textit{global} level\textemdash jointly forecasting one future for all participants at
a time, rather than at a \textit{local} output level\textemdash one future for each individual independent of the remaining participants' futures. In contrast, applying the prior formulations \cite{ahujaReactNotReact2019, jooSocialArtificialIntelligence2019, palmero2022chalearn} requires the training of separate models treating each individual as a target (for the same group input) and then forecasting an independent future one at a time. Meanwhile, other prior pose forecasting works \cite{chaoForecastingHumanDynamics2017,fragkiadakiRecurrentNetworkModels2015, walkerPoseKnowsVideo2017, habibieRecurrentVariationalAutoencoder2017, pavlloQuaterNetQuaternionbasedRecurrent2018} have been in non-social settings and do not need to model such behavioral interdependence.

\paragraph{Non-Contiguous Observed and Future Windows.} Domain experts are often interested in settings where $\tobs$ and $\tfut$ are offset by an arbitrary delay, such as forecasting a time lagged synchrony \cite{delahercheInterpersonalSynchronySurvey2012} or mimicry \cite{bilakhiaAudiovisualDetectionBehavioural2013} episode, or upcoming disengagement \cite{bohusManagingHumanRobotEngagement2014, vandoornRitualsLeavingPredictive2018}. We therefore allow for non-contiguous $\tobs$ and $\tfut$. Operationalizing prior formulations that predict one step into the future \cite{yaoMultipleGranularityGroup2018, sanghvi2019mgpi, jooSocialArtificialIntelligence2019, airaleSocialInteractionGANMultipersonInteraction2021} would entail a sliding window of autoregressive predictions over the offset between $\tobs$ and $\tfut$ (from $oT$ to $f1$), with errors cascading even before decoding is performed over the window of interest $\tfut$.

\medskip
\noindent Our task formalization of SCF can be viewed as a social science-grounded generalization of prior computational formulations, and therefore suitable for a wider range of cross-disciplinary tasks, both computational and analytical.

\section{Method Preliminaries} \label{sec:background}

\paragraph{Meta-Learning.} A supervised learning algorithm can be viewed as a function mapping a dataset $C \coloneqq (\bm{X}_C, \bm{Y}_C) \coloneqq \{(\bm{x}^i, \bm{y}^i)\}_{i \in [N_C]}$ to a predictor $f(\bm{x})$. Here $N_C$ is the number of datapoints in $C$, and $[N_C] \coloneqq \{1, \ldots, N_C\}$. The key idea of meta-learning is to learn how to learn from a dataset in order to adapt to unseen supervised tasks; hence the name \textit{meta}-learning. This is done by learning a map $C \mapsto f(\cdot,C)$.
In meta-learning literature, a \textit{task} refers to each dataset in a collection $\{\mathcal{T}_m\}_{m=1}^{N_{\mathrm{tasks}}}$ of related datasets \cite{hospedalesMetaLearningNeuralNetworks2020}. Training is episodic, where each task $\mathcal{T}$ is split into subsets $(C, D)$. A meta-learner then fits the subset of target points $D$ given the subset of context observations $C$. At meta-test time, the resulting predictor $f(\bm{x},C)$ is adapted to make predictions for target points on  an unseen task by conditioning on a new context set $C$ unseen during meta-training.

\paragraph{Neural Processes (NPs).} Sharing the same core motivations, NPs \cite{garneloNeuralProcesses2018} can be viewed as a family of latent variable models that extend the idea of meta-learning to situations where uncertainty in the predictions $f(\bm{x},C)$ are desirable. They do this by meta-learning a map from datasets to stochastic processes, estimating a distribution over the predictions $p(\bm{Y}|\bm{X},C)$. To capture this distribution, NPs model the conditional latent distribution $p(\bm{z}|C)$ from which a task representation $\bm{z} \in \mathbb{R}^d$ is sampled. This introduces stochasticity, constituting what is called the model's \textit{latent path}. The context can also be directly incorporated through a \textit{deterministic path}, via a representation $\bm{r}_C \in \mathbb{R}^d$ aggregated over $C$. An observation model $p(\bm{y}^i|\bm{x}^i,\bm{r}_C,\bm{z})$ then fits the target observations in $D$.
The generative process for the NP is written as
\begin{equation}\label{eq:np}
    p(\bm{Y}|\bm{X},C) \coloneqq \int p(\bm{Y}|\bm{X},C,\bm{z})p(\bm{z}|C)d\bm{z}
    = \int p(\bm{Y}|\bm{X},\bm{r}_C,\bm{z})q(\bm{z}|\bm{s}_C)d\bm{z},
\end{equation}
where  $p(\bm{Y}|\bm{X},\bm{r}_C,\bm{z}) \coloneqq \prod_{i \in [N_D]} p(\bm{y}^i|\bm{x}^i,\bm{r}_C,\bm{z})$. The latent $\bm{z}$ is modeled by a factorized Gaussian parameterized by $\bm{s}_C \coloneqq f_s(C)$, with $f_s$ being a deterministic function invariant to order permutation over $C$. When the conditioning on context is removed $(C = \emptyset$), we have $q(\bm{z}|\bm{s}_\emptyset) \coloneqq p(\bm{z})$, the zero-information prior on $\bm{z}$. The deterministic path uses a function $f_r$ similar to $f_s$, so that $\bm{r}_C \coloneqq f_r(C)$. In practice this is implemented as $\bm{r}_C = \sum_{i \in [N_C]}\mathrm{MLP}(\bm{x}_i, \bm{y}_i) / N_C$.
The observation model is referred to as the \textit{decoder}, and
$q, f_r, f_s$ comprise the \textit{encoders}. The parameters of the NP are learned for random subsets $C$ and $D$ for a task by maximizing the evidence lower bound (ELBO)
\begin{equation}
     \log p(\bm{Y}|\bm{X},C) \geq \mathbb{E}_{q(\bm{z}|\bm{s}_D)}[\log p(\bm{Y}|\bm{X},C,\bm{z})] - \mathbb{KL}(q(\bm{z}|\bm{s}_D) || q(\bm{z}|\bm{s}_C)). \label{eq:elbo}
\end{equation}

\section{Social Processes: Methodology} \label{sec:methodology}
Our core idea for adapting predictions to a group’s unique behavioral dynamics is to condition forecasts on a context set $C$ of the same group's observed-future sequence pairs.
By \textit{learning to learn}, i.e., \textit{meta-learn} from a context set, our model can generalize to unseen groups at evaluation by conditioning on an unseen context set of the test group's behavior sequences. In practice, a social robot might, for instance, observe such an evaluation context set before approaching a new group.

We set up by splitting the interaction into pairs of observed and future sequences, writing the context
as $C \coloneqq (\bm{X}_C, \bm{Y}_C) \coloneqq (\bm{X}_j, \bm{Y}_k)_{(j, k) \in [N_C] \times [N_C]}$, where every $\bm{X}_j$ occurs before the corresponding $\bm{Y}_k$. Since we allow for non-contiguous $\tobs$ and $\tfut$, the $j$th $\tobs$ can have multiple associated $\tfut$ windows for prediction, up to a maximum offset. Denoting the set of target window pairs as $D \coloneqq (\bm{X}, \bm{Y}) \coloneqq (\bm{X}_j, \bm{Y}_k)_{(j, k) \in [N_D]\times[N_D]}$, our goal is to model the distribution $p(\bm{Y}|\bm{X},C)$. Note that when conditioning on context is removed ($C = \emptyset$), we simply revert to the non-meta-learning formulation $p(\bm{Y}|\bm{X})$.

\begin{figure}[!t]
\makebox[\textwidth][c]{\includegraphics[width=\textwidth]{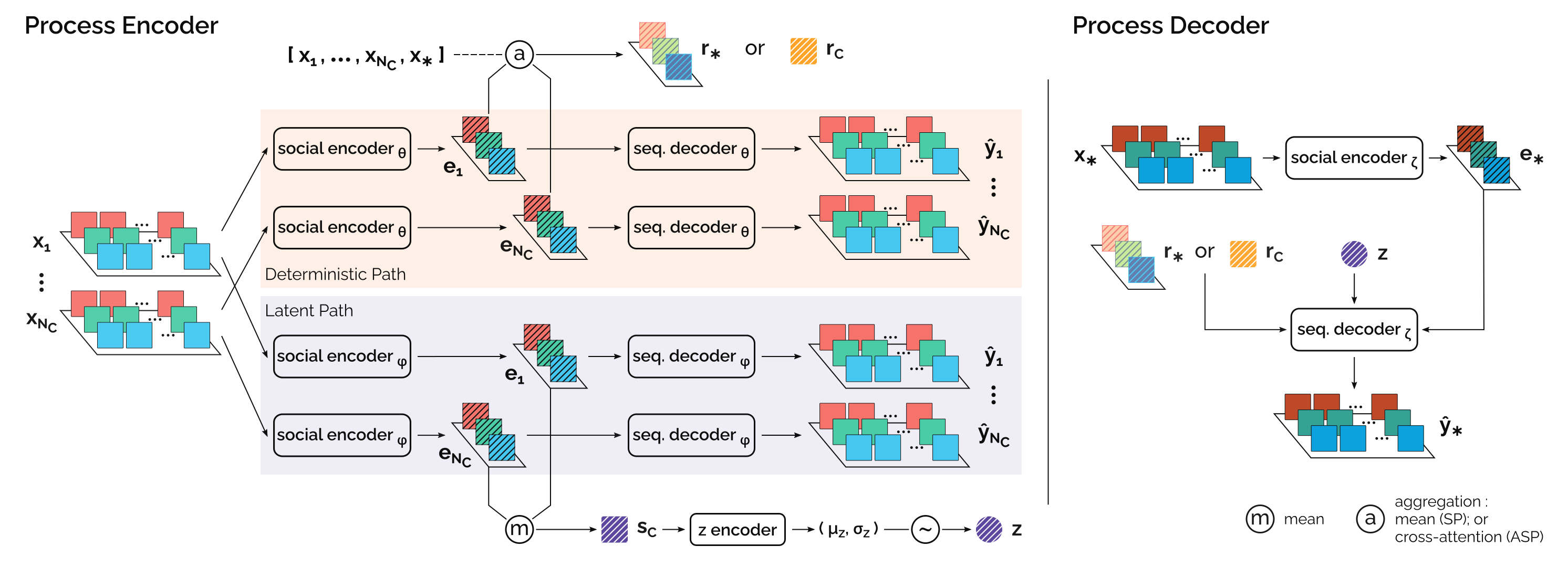}}
\caption{Architecture of the SP and ASP family.}
\label{fig:architecture}
\end{figure}

The generative process for our Social Process (SP) model follows \autoref{eq:np}, which we extend to social forecasting in two ways. We embed an observed sequence $\bm{x}^i$ for participant $\mathrm{p}_i$ into a condensed encoding $\bm{e^i} \in \mathbb{R}^{d}$ that is then decoded into the future sequence using a Seq2Seq architecture \cite{sutskeverSequenceSequenceLearning2014, choLearningPhraseRepresentations2014}. Crucially, the sequence decoder only accesses $\bm{x}^i$ through $\bm{e}^i$. So after training, $\bm{e}^i$ must encode the \textit{temporal} information that $\bm{x}^i$ contains about the future. Further, social behavior is interdependent. We model $\bm{e}^i$ as a function of both, $\mathrm{p}_i$'s own behavior as well as that of partners $\mathrm{p}_{j, j\neq i}$ from $\mathrm{p}_i$'s perspective. This captures the \textit{spatial} influence partners have on the participant over $\tobs$. Using notation we established in \Cref{sec:task}, we define the observation model for $\mathrm{p}_i$ as
\begin{equation}
    p(\bm{y}^i|\bm{x}^i,C,\bm{z})
    \coloneqq p(\bm{b}_{f1}^i,\ldots,\bm{b}_{fT}^i | \bm{b}_{o1}^i,\ldots,\bm{b}_{oT}^i,C,\bm{z})
    = p(\bm{b}_{f1}^i,\ldots,\bm{b}_{fT}^i | \bm{e}^i,\bm{r}_C,\bm{z}).\label{eq:spdef}
\end{equation}
If decoding is carried out in an auto-regressive manner, the right hand side of \autoref{eq:spdef} simplifies to $\prod_{t=f1}^{fT} p(\bm{b}_{t}^i|\bm{b}_{t-1}^i, \ldots, \bm{b}_{f1}^i, \bm{e}^i,\bm{r}_C,\bm{z})$. Following the standard NP setting, we implement the observation model as a set of Gaussian distributions factorized over time and feature dimensions. We also incorporate the cross-attention mechanism from the Attentive Neural Process (ANP) \cite{kimAttentiveNeuralProcesses2019} to define the variant Attentive Social Process (ASP). Following \autoref{eq:spdef} and the definition of the ANP, the corresponding observation model of the ASP for a single participant is defined as
\begin{equation}
    p(\bm{y}^i|\bm{x}^i,C,\bm{z}) = p(\bm{b}_{f1}^i,\ldots,\bm{b}_{fT}^i|\bm{e}^i,r^*(C, \bm{x}^i),\bm{z}). \label{eq:aspdef}
\end{equation}
Here each target query sequence $\bm{x}_*^i$ attends to the context sequences $\bm{X}_C$ to produce a query-specific representation $\bm{r}_* \coloneqq r^*(C, \bm{x}_*^i) \in \mathbb{R}^{d}$.

The model architectures are illustrated in \autoref{fig:architecture}. Note that our modeling assumption is that the underlying stochastic process generating social behaviors does not evolve over time. That is, the individual factors determining how participants coordinate behaviors\textemdash age, cultural background, personality variables \citetext{\citealp[Chap.~1]{moore2013stacks}; \citealp[p.~237]{kendonConductingInteractionPatterns1990}}\textemdash are likely to remain the same over a single interaction. This is in contrast to the line of work that deals with \textit{meta-transfer learning}, where the stochastic process itself changes over time \cite{singhSequentialNeuralProcesses2019, yoonRobustifyingSequentialNeural2020, williRecurrentNeuralProcesses2019, kumarSpatiotemporalModelingUsing}; this entails modeling a different $\bm{z}$ distribution for every timestep.

\paragraph{Encoding Partner Behavior.} To encode partners' influence on an individual's future, we use a pair of sequence encoders: one to encode the temporal dynamics of participant $\mathrm{p}^i$'s features, $\bm{e}_\mathrm{self}^i = f_\mathrm{self}(\bm{x}^i)$, and another to encode the dynamics of a transformed representation of the features of $\mathrm{p}^i$'s partners, $\bm{e}_\mathrm{partner}^i = f_\mathrm{partner}(\psi(\bm{x}^{j, (j \neq i)}))$. Using a separate network to encode partner behavior enables sampling an individual's and partners' features at different sampling rates.

\begin{figure}[!t]
\makebox[\textwidth][c]{\includegraphics[width=0.9\textwidth]{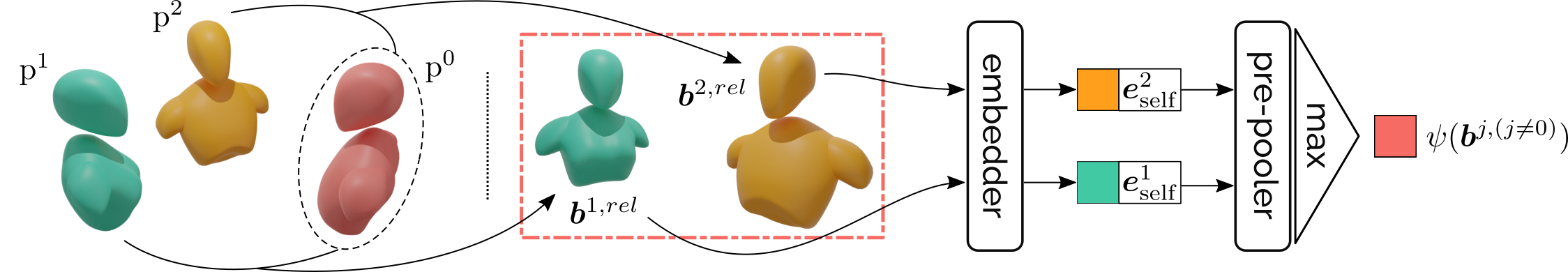}}
\caption{Encoding partner behavior for participant $\mathrm{p}^0$ for a single timestep. To model the influence partners  $\mathrm{p}^1$ and $\mathrm{p}^2$ have on the behavior of $\mathrm{p}^0$, we transform the partner features to capture the interaction from $\mathrm{p}^0$'s perspective, and learn a representation of these features invariant to group size and partner-order permutation using the symmetric $\mathrm{max}$ function.}
\label{fig:pooling}
\end{figure}

How do we model $\psi(\bm{x}^{j, (j \neq i)})$? We want the partners' representation to possess two properties: \textit{permutation invariance}\textemdash changing the order of the partners should not affect the representation, and \textit{group-size independence}\textemdash we want to compactly represent all partners independent of the group size. Intuitively, to model partner influence on $\mathrm{p}^i$, we wish to \textit{capture a view of the partners' behavior as $\mathrm{p}^i$ perceives it}. Figure~\ref{fig:pooling} illustrates the underlying intuition. We do this by computing pooled embeddings of relative behavioral features, extending \citet{guptaSocialGANSocially2018}'s approach for pedestrian positions to conversation behavior. Note that our partner-encoding approach is in contrast to that of \citet{tanMultimodalJointHead2021}, which is order and group-size dependent, and \citet{yaoMultipleGranularityGroup2018}, who do not transform the partner features to an individual's perspective.

Since the most commonly considered cues in literature are pose (orientation and location) and binary speaking status \cite{alameda-pinedaAnalyzingFreestandingConversational2015, zhangSocialInvolvementMingling2018, tanMultimodalJointHead2021}, we specify how we transform them. For a single timestep, we denote these cues for $\mathrm{p}^i$ as $\bm{b}^i = [\bm{q}^i;\bm{l}^i;s^i]$, and for $\mathrm{p}^j$ as $\bm{b}^j = [\bm{q}^j;\bm{l}^j;s^j]$. We compute the relative partner features $\bm{b}^{j, rel} = [\bm{q}^{rel};\bm{l}^{rel};s^{rel}]$ by transforming $\bm{b}^j$ to a frame of reference defined by $\bm{b}^i$:
\begin{subequations}
\begin{gather}
    \bm{q}^{rel} = \bm{q}^i * (\bm{q}^j)^{-1}, \quad
    \bm{l}^{rel} = \bm{l}^j - \bm{l}^i,  \quad
    s^{rel} = s^j - s^i. \tag{\theequation a-c}
\end{gather}
\end{subequations}
Note that we use unit quaternions (denoted $\bm{q}$) for representing orientation due to their various benefits over other representations of rotation \citep[Sec.~3.2]{kendallGeometricLossFunctions2017}. The operator $*$ denotes the Hamilton product of the quaternions. These transformed features $\bm{b}^{j, rel}$ for each $\mathrm{p}^j$ are then encoded using an \textit{embedder} MLP. The outputs are concatenated with their corresponding $\bm{e}_\mathrm{self}^j$ and processed by a \textit{pre-pooler} MLP. Assuming  $d_\mathrm{in}$ and $d_\mathrm{out}$ pre-pooler input and output dims and $J$ partners, we stack the $J$ inputs to obtain $(J, d_\mathrm{in})$ tensors. The $(J, d_\mathrm{out})$-dim output is element-wise max-pooled over the $J$ dim, resulting in the $d_\mathrm{out}$-dim vector $\psi(\bm{b}^{j, (j \neq i)})$ for any value of $J$, per timestep. We capture the temporal dynamics in this pooled representation over $\tobs$ using $f_\mathrm{partner}$. Finally, we combine $\bm{e}_\mathrm{self}^i$ and $\bm{e}_\mathrm{partner}^i$ for $\mathrm{p}^i$ through a linear projection (defined by a weight matrix $W$) to obtain the individual's embedding $\bm{e}_\mathrm{ind}^i = W\cdot [\bm{e}_\mathrm{self}^i;\bm{e}_\mathrm{partner}^i]$. Our intuition is that with information about both $\mathrm{p}^i$ themselves, and of $\mathrm{p}^i$'s partners from $\mathrm{p}^i$'s point-of-view,  $\bm{e}_\mathrm{ind}^i$ now contains the information required to predict $\mathrm{p}^i$'s future behavior.

\paragraph{Encoding Future Window Offset.} Since we allow for non-contiguous windows, a single $\tobs$ might be associated to multiple $\tfut$ windows at different offsets. Decoding the same $\bm{e}_\mathrm{ind}^i$ into multiple sequences (for different $\tfut$) in the absence of any timing information might cause an averaging effect in either the decoder or the information encoded in $\bm{e}_\mathrm{ind}^i$. One option would be to immediately start decoding after $\tobs$ and discard the predictions in the offset between $\tobs$ and $\tfut$. However, auto-regressive decoding might lead to cascading errors over the offset. Instead, we address this one-to-many issue by injecting the offset information into $\bm{e}_\mathrm{ind}^i$. The decoder then receives a unique encoded representation for every $\tfut$ corresponding to the same $\tobs$. We do this by repurposing the idea of sinusoidal positional encodings \cite{vaswaniAttentionAllYou2017} to encode window offsets rather than relative token positions in sequences. For a given $\tobs$ and $\tfut$, and $d_e$-dim $\bm{e}_\mathrm{ind}^i$ we define the offset as $\Delta t = f1 - oT$, and the corresponding offset encoding $OE_{\Delta t}$ as
\begin{subequations}
\begin{gather}
    OE_{(\Delta t, 2m)} = \sin(\Delta t/10000^{2m/d_e}),
    OE_{(\Delta t, 2m+1)} = \cos(\Delta t/10000^{2m/d_e}). \tag{\theequation a, b}
\end{gather}
\end{subequations}
Here $m$ refers to the dimension index in the encoding. We finally compute the representation $\bm{e}^i$ for \autoref{eq:spdef} and \autoref{eq:aspdef} as
\begin{equation}
    \bm{e}^i = \bm{e}_\mathrm{ind}^i + OE_{\Delta t}.
\end{equation}

\paragraph{Auxiliary Loss Functions.} We incorporate a geometric loss function for each of our sequence decoders to improve performance in pose regression tasks. For $\mathrm{p}_i$ at time $t$, given the ground truth $\bm{b}^i_t = [\bm{q};\bm{l};s]$, and
the predicted mean $\bm{\hat{b}}^i_t = [\bm{\hat{\bm{q}}};\bm{\hat{\bm{l}}};\hat{s}]$, we denote the tuple $(\bm{b}^i_t, \bm{\hat{b}}^i_t)$ as $B^i_t$. We then have the location loss in Euclidean space $\mathcal{L}_{\mathrm{l}}(B^i_t) = || \bm{l} - \bm{\hat{\bm{l}}} ||$, and we can regress the quaternion values using
\begin{equation}
    \mathcal{L}_{\mathrm{q}}(B^i_t) = \norm{\bm{q} - \frac{\hat{\bm{q}}}{\norm{\hat{\bm{q}}}}}.
\end{equation}
\citet{kendallGeometricLossFunctions2017} show how these losses can be combined using the homoscedastic uncertainties in position and orientation, $\hat{\sigma}^{2}_{\mathrm{l}}$ and $\hat{\sigma}^{2}_{\mathrm{q}}$:
\begin{equation}
    \mathcal{L}_{\sigma}(B^i_t) = \mathcal{L}_{\mathrm{l}}(B^i_t)\exp(-\hat{s}_{\mathrm{l}}) + \hat{s}_{\mathrm{l}}+\mathcal{L}_{\mathrm{q}}(B^i_t)\exp(-\hat{s}_{\mathrm{q}}) + \hat{s}_{\mathrm{q}},
\end{equation}
where $\hat{s} \coloneqq \log\hat{\sigma}^2$. Using the binary cross-entropy loss for speaking status $\mathcal{L}_{\mathrm{s}}(B^i_t)$, we have the overall auxiliary loss over $t \in \tfut$:
\begin{equation}
    \mathcal{L}_{\mathrm{aux}}(\bm{Y}, \bm{\hat{Y}}) = \sum_i \sum_t \mathcal{L}_{\sigma}(B^i_t) + \mathcal{L}_{\mathrm{s}}(B^i_t).
\end{equation}
The parameters of the SP and ASP are trained by maximizing the ELBO (\autoref{eq:elbo})
and minimizing this auxiliary loss.

\section{Experiments and Results}
\label{sec:experiments}
\subsection{Experimental Setup}

\paragraph{Evaluation Metrics.} Prior forecasting formulations output a single future. However, since the future is not deterministic, we predict a future distribution. Consequently, needing a metric that accounts for probabilistic predictions, we report the log-likelihood (LL) $\log p(\bm{Y}|\bm{X},C)$, commonly used by all variants within the NP family \cite{garneloNeuralProcesses2018, kimAttentiveNeuralProcesses2019, singhSequentialNeuralProcesses2019}. The metric is equal to the log of the predicted density evaluated at the ground-truth value. (Note: the fact that the vast majority of forecasting works even in pedestrian settings omit a probabilistic metric, using only geometric metrics, is a limitation also observed by \citet[Sec.~8.3]{rudenko2020human}.) Nevertheless, for additional insight beyond the LL, we also report the errors in the predicted means\textemdash geometric errors for pose and accuracy for speaking status\textemdash and provide qualitative visualizations of forecasts.

\paragraph{Models and Baselines.} In keeping with the task requirements and for fair evaluation, we require that all models we compare against forecast a distribution over future cues. 
\begin{itemize}[itemsep=0em]
\item To evaluate our core idea of viewing conversing groups as meta-learning tasks, we compare against non-meta-learning methods: we adapt variational encoder-decoder (VED) architectures \cite{haNeuralRepresentationSketch2017, bowmanGeneratingSentencesContinuous2016} to output a distribution. 
\item To evaluate our specific modeling choices within the meta-learning family, we compare against the NP and ANP models (see Section~\ref{sec:methodology}). The original methods were not proposed for sequences, so we adapt them by collapsing the timestep and feature dimensions in the data.
\end{itemize}
Note that in contrast to the SP models, these baselines have direct access to the future sequences in the context, and therefore constitute a strong baseline.
We consider two variants for both NP and SP models: \textit{-latent} denoting only the stochastic path; and \textit{-uniform} containing both the deterministic and stochastic paths with uniform attention over context sequences. We further consider two attention mechanisms for the cross-attention module: \textit{-dot} with dot attention, and \textit{-mh} with wide multi-head attention \cite{kimAttentiveNeuralProcesses2019}.
Finally, we experiment with two choices of backbone architectures: multi-layer perceptrons (MLP), and Gated Recurrent Units (GRU). Implementation and training details can be found in Appendix~\ref{app:implementation}. Code, processed data, trained models, and test batches for reproduction are available at {\small \url{https://github.com/chiragraman/social-processes}}.

\subsection{Evaluation on Synthesized Behavior Data}
\begin{figure}[!t]
\begin{minipage}[t]{0.56\textwidth}
  \centering
  \captionsetup{type=figure}
  \includegraphics[width=\textwidth]{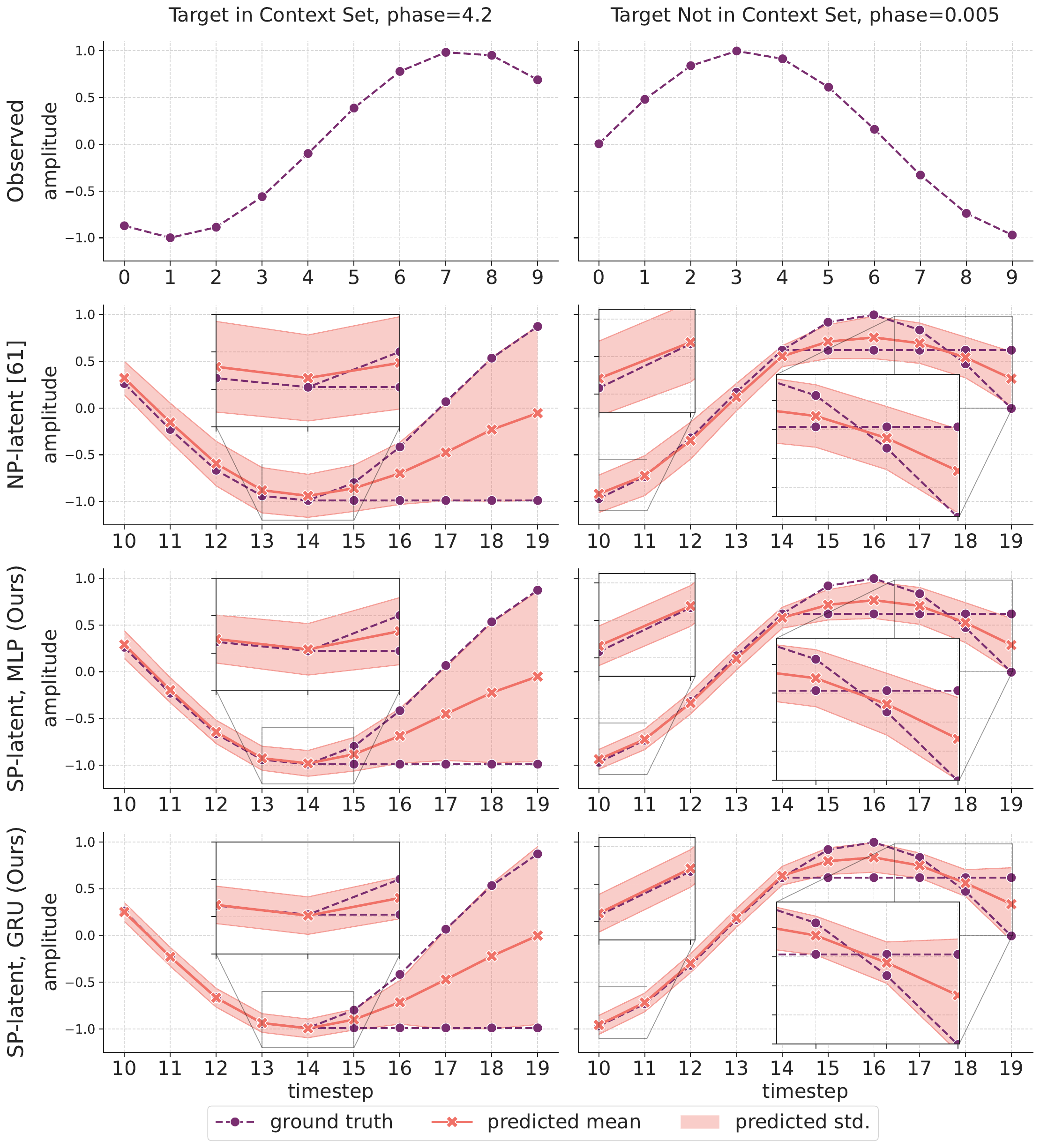}
  \captionof{figure}{Ground truths and predictions for the toy task of forecasting simulated glancing behavior. Our SP models learn a better fit than the NP model, SP-GRU being the best (see zoomed insets).}
  \label{fig:toy}
\end{minipage}
\hfill
\begin{minipage}[t]{0.43\textwidth}
  \centering
  \captionsetup{type=figure}
  \includegraphics[width=\textwidth]{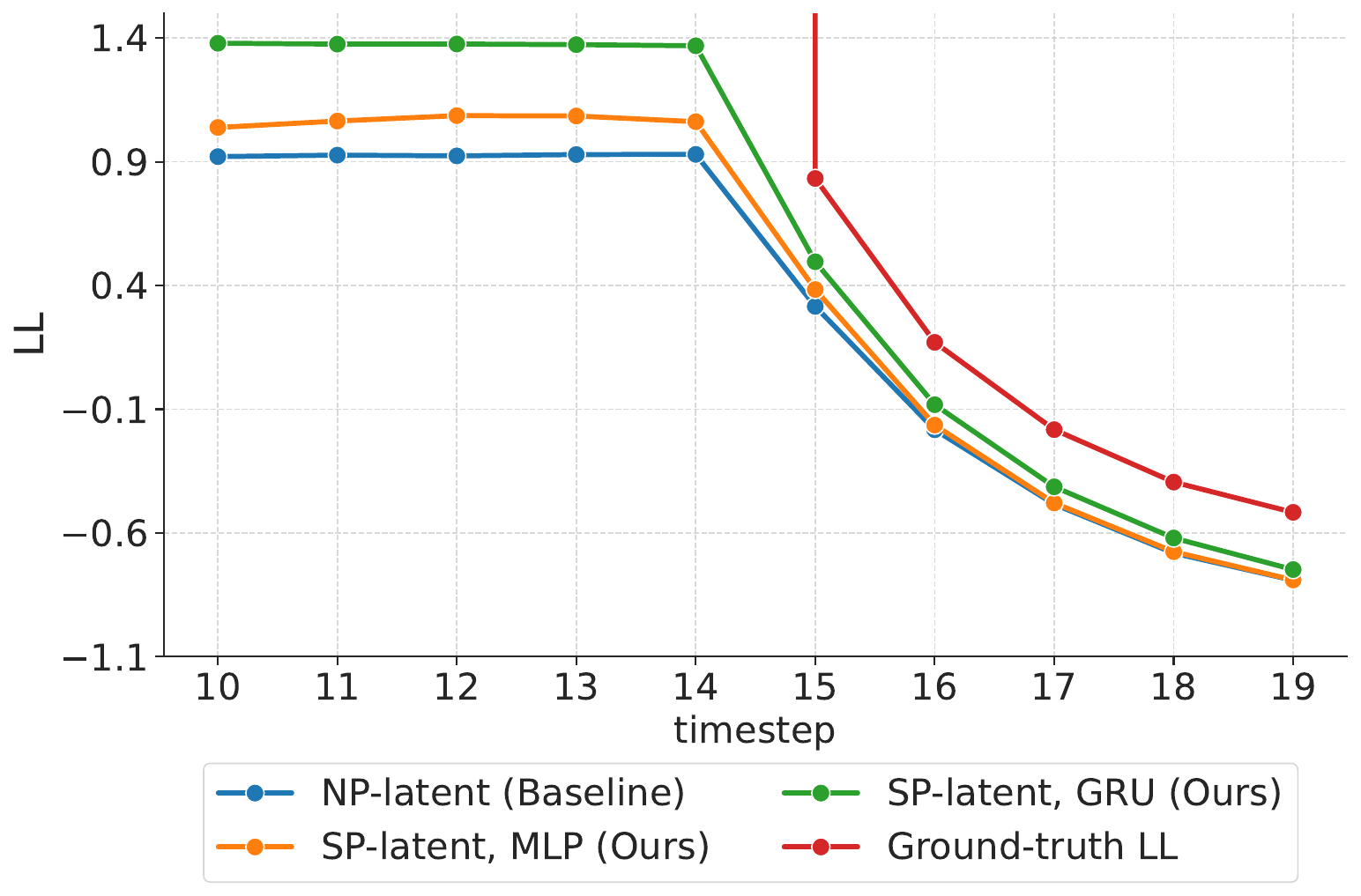}
  \captionof{figure}{Mean per timestep LL over the sequences in the synthetic glancing dataset. Higher is better.}
  \label{fig:ts-synthetic-ll}
  \captionsetup{type=table}
  \ra{1.1}
\setlength{\tabcolsep}{8.5pt}
\centering
\scriptsize
\captionof{table}{\textbf{Mean (Std.) Metrics on the Synthetic Glancing Behavior Dataset.} The metrics are averaged over timesteps; mean and std. are then computed over sequences. Higher is better for LL, lower for MAE.}
\label{tab:glancing-metrics}

\begin{tabular}[b]{@{}lcc@{}}
\toprule
  & \multirow{2}{*}{\textbf{LL}} & \textbf{Head Ori.} \\
  & & MAE (\degree) \\

\midrule
  NP-latent & 0.28~(0.24) & 19.63~(7.26)  \\
  
\specialrule{\lightrulewidth}{1ex}{1ex}
  SP-latent (MLP)  & 0.36~(0.20) & 19.46~(7.05) \\
  SP-latent (GRU)  & \textbf{0.55~(0.23)} & \textbf{18.55~(7.11)} \\
\bottomrule
\end{tabular}

\end{minipage}
\vspace{-2pt}
\end{figure}

To first validate our method on a toy task,
we synthesize a dataset simulating two glancing behaviors in social settings \cite{mooreNonverbalCourtshipPatterns1985}, approximated by horizontal head rotation. The sweeping \textit{Type I} glance is represented by a 1D sinusoid over $20$ timesteps. The gaze-fixating \textit{Type III} glance is denoted by clipping the amplitude for the last six timesteps. The task is to forecast the signal over the last $10$ timesteps ($\tfut$) by observing the first $10$ ($\tobs$). Consequently, the first half of $\tfut$ is certain, while the last half is uncertain: every observed sinusoid has two ground truth futures in the data (clipped and unclipped). It is impossible to infer from an observed sequence alone if the head rotation will stop partway through the future.
\autoref{fig:toy} illustrates the predictions for two sample sequences. Table~\ref{tab:glancing-metrics} provides quantitative metrics and \autoref{fig:ts-synthetic-ll} plots the LL per timestep. The LL is expected to decrease over timesteps where ground-truth futures diverge, being $\infty$ when the future is certain. We observe that all models estimate the mean reasonably well, although our proposed SP models perform best. More crucially, the SP models, especially the SP-GRU, learn much better uncertainty estimates compared to the NP baseline (see zoomed regions in \autoref{fig:toy}). We provide additional analysis, alternative qualitative visualizations, and data synthesis details in \Cref{app:results,app:viz,app:data} respectively.

\subsection{Evaluation on Real-World Behavior Data}  \label{subsec:dataset}

\paragraph{Datasets and Preprocessing.} With limited behavioral data availability, a common practice in the domain is to solely train and evaluate methods on synthesized behavior dynamics \cite{vazquezMaintainingAwarenessFocus2016,sanghvi2019mgpi}. In contrast, we also evaluate on two real-world behavior datasets: the MatchNMingle (MnM) dataset of in-the-wild mingling behavior \cite{cabrera2018matchnmingle}, and the Haggling dataset of a triadic game where two sellers compete to sell a fictional product to a buyer \cite{jooSocialArtificialIntelligence2019}.
For MnM, we treat the $42$ groups from Day 1 as test sets and a total of $101$ groups from the other two days as train sets. For Haggling,
we use the same split of $79$ training and $28$ test groups used by \citet{jooSocialArtificialIntelligence2019}.
We consider the following cues: \textit{head pose} and \textit{body pose}, described by the location of a keypoint and an orientation quaternion; and binary \textit{speaking status}. These are the most commonly considered cues in computational analyses of conversations \cite{alameda-pinedaAnalyzingFreestandingConversational2015, zhangSocialInvolvementMingling2018, tanMultimodalJointHead2021} given how crucial they are in sustaining interactions \cite{kendonConductingInteractionPatterns1990, duncanSignalsRulesTaking1972, vinciarelliSocialSignalProcessing2009a}. For orientation, we first convert the normal vectors (provided in the horizontal direction in both datasets) into unit quaternions. Since the quaternions $\mathbf{q}$ and $-\mathbf{q}$ denote an identical rotation, we constrain the first quaternion in every sequence to the same hemisphere and interpolate subsequent quaternions to have the shortest distance along the unit hypersphere.
We then split the interaction data into pairs of $\tobs$ and $\tfut$ windows to construct the samples for forecasting. We specify dataset-specific preprocessing details in Appendix~\ref{app:data}.

\begin{figure}[!t]
\begin{minipage}[t]{0.52\textwidth}
  \captionsetup{type=table}
  \centering
\scriptsize
\ra{1.14}
\setlength{\tabcolsep}{0.5pt}
\captionof{table}{\textbf{Mean (Std.) Log-Likelihood (LL) on the MatchNMingle and Haggling Test Sets.} For a single sequence, we sum over the feature and participant dimensions, and average over timesteps. The reported mean and std. are over individual sequences in the test sets. Higher is better. Underline indicates best LL within family.}
\label{tab:nll}

\begin{tabular}[b]{@{}lcccc@{}}
\toprule
  & \multicolumn{2}{c}{\textbf{MatchNMingle}} & \multicolumn{2}{c}{\textbf{Haggling}} \\
  \cmidrule{2-5}
  & \textbf{Random} & \textbf{Fixed-Initial} & \textbf{Random} & \textbf{Fixed-Initial} \\

\midrule
  \multicolumn{2}{@{}l}{\textbf{VED Family \cite{haNeuralRepresentationSketch2017, bowmanGeneratingSentencesContinuous2016}}} \\
  VED-MLP   & 8.1~(7.2) & 7.9~(7.0) & 4.0~(8.3) & 4.1~(8.2) \\
  VED-GRU   & 25.4~(18.0) & 25.1~(19.1) & 60.3~(2.2) & 60.3~(2.1) \\
  
\specialrule{\lightrulewidth}{1ex}{1ex}
  \multicolumn{2}{@{}l}{\textbf{NP Family \cite{garneloNeuralProcesses2018, kimAttentiveNeuralProcesses2019}}} \\
  NP-latent     & 22.1~(17.8) & \underline{21.6}~(18.5) & \underline{27.2}~(17.3) & \underline{27.9}~(16.3) \\
  NP-uniform    & 21.4~(18.8) & 20.5~(17.8) & 24.8~(22.9) & 25.0~(22.2) \\
  ANP-dot       & 22.8~(18.6) & 21.0~(18.3) & 26.7~(21.4) & 24.7~(20.8) \\
  ANP-mh        & \underline{23.6}~(15.6) & 20.0~(23.9) & 25.1~(23.1) & 24.8~(22.4) \\

\specialrule{\lightrulewidth}{1ex}{1ex}
  \multicolumn{2}{@{}l}{\textbf{Ours (SP-MLP)}} \\
  SP-latent     & 102.1~(29.9) & 101.5~(29.2) & 136.6~(7.0) & 136.7~(7.0) \\
  SP-uniform    & 112.8~(34.1) & \underline{111.4}~(33.8) & 138.3~(8.0) & 137.6~(8.4) \\
  ASP-dot       & 109.9~(32.9) & 107.6~(32.1) & 137.8~(7.5) & 136.4~(7.6) \\
  ASP-mh        & \underline{112.9}~(34.7) & 111.3~(33.6) & \underline{146.0}~(10.9) & \underline{145.7}~(10.2) \\ [1ex]

  \multicolumn{2}{@{}l}{\textbf{Ours (SP-GRU)}} \\
  SP-latent     & 86.4~(37.2) & 85.4~(37.2) & 66.7~(27.4) & 66.2~(30.7) \\
  SP-uniform    & 87.0~(38.4) & \underline{85.5}~(38.3) & \underline{79.9}~(50.5) & \underline{78.6}~(52.2) \\
  ASP-dot       & \underline{87.6}~(39.1) & 83.9~(38.1) & 38.4~(60.4) & 27.2~(93.4) \\
  ASP-mh        & 85.8~(37.1) & 82.3~(36.0) & 66.3~(30.3) & 59.3~(32.4) \\
\bottomrule
\end{tabular}

\end{minipage}
\hfill
\begin{minipage}[t]{0.455\textwidth}
  \centering
  \captionsetup{type=figure}
  \includegraphics[width=\textwidth]{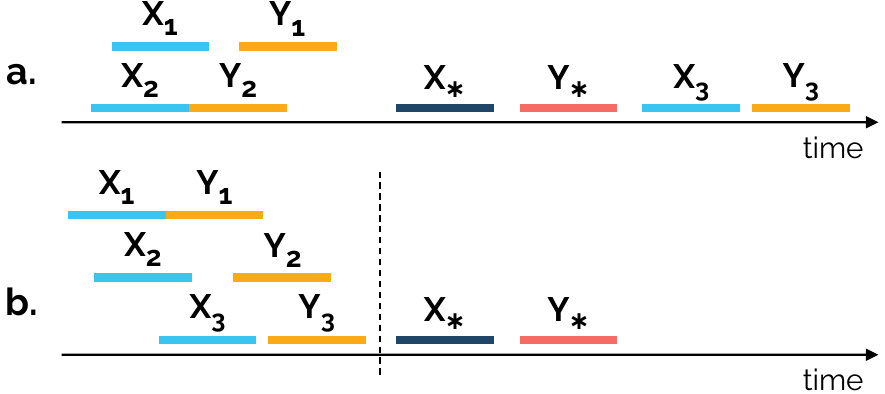}
  \captionof{figure}{Context Regimes. For a target sequence pair ($\bm{X}_*$, $\bm{Y}_*$), context pairs (here $3$) are sampled either \textbf{a.} randomly across the lifetime of the group interaction (\textit{random}), or \textbf{b.} from a fixed initial duration (\textit{fixed-initial}).}
  \label{fig:contexts}
  \vfill
  \centering
  \captionsetup{type=figure}
  \includegraphics[width=\textwidth]{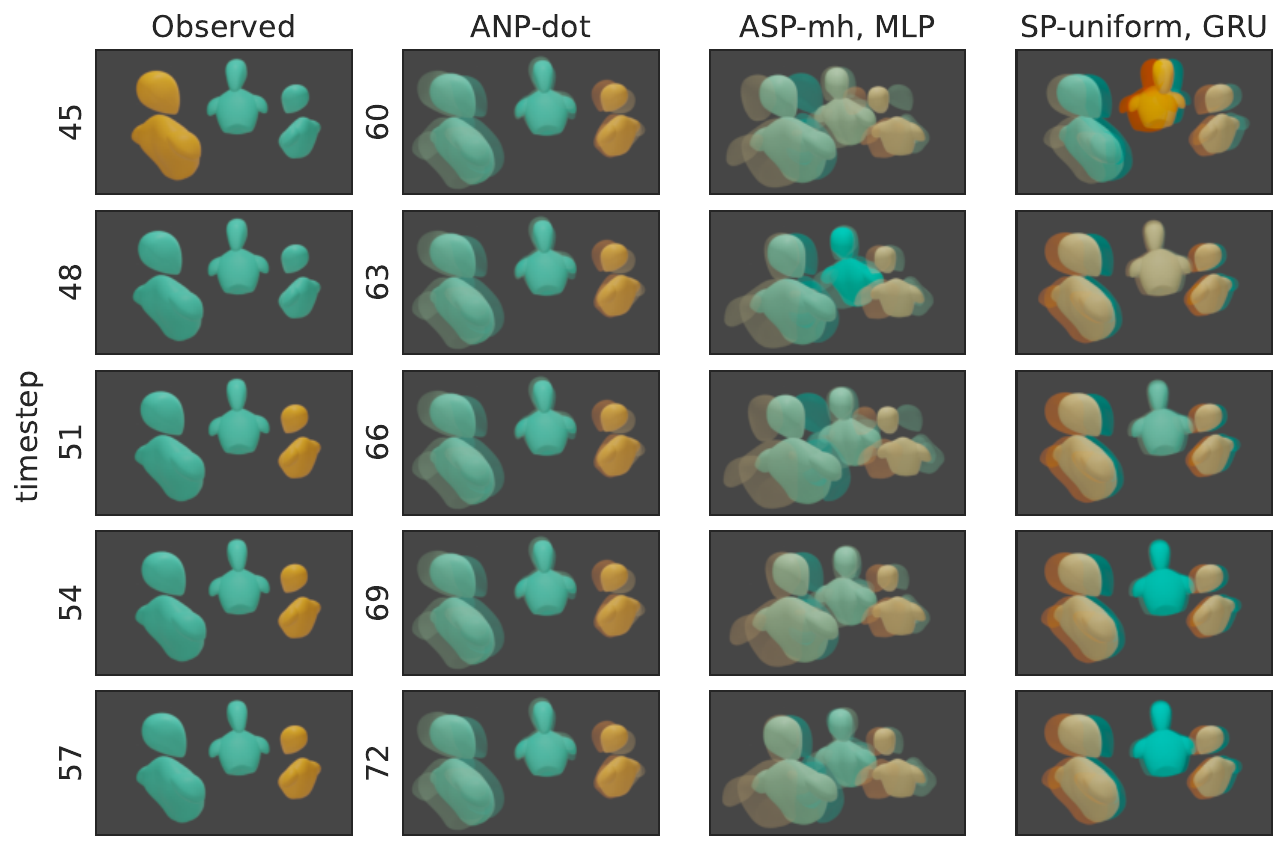}
  \captionof{figure}{Forecasts over selected timesteps from the Haggling group \textit{170224-a1-group1}. Speaking status is interpolated between orange (speaking) and blue (listening). Translucent models denote the predicted $\text{mean} \pm \text{std.}$}
  \label{fig:haggling-train-reduced}
\end{minipage}
\end{figure}

\paragraph{Context Regimes.} We evaluate on two context regimes: \textit{random}, and \textit{fixed-initial} (see \autoref{fig:contexts}). In the \textit{random} regime, context samples (observed-future pairs) are selected as a random subset of target samples, so the model is exposed to behaviors from any phase of the interaction lifecycle. Here we ensure that batches contain unique $\tobs$ to prevent any single observed sequence from dominating the aggregation of representations over the context split. At evaluation, we take $50\%$ of the batch as context. The \textit{fixed-initial} regime investigates how models can learn from observing the initial dynamics of an interaction where certain gestures and patterns are more distinctive \citep[Chap.~6]{kendonConductingInteractionPatterns1990}. Here we treat the first $20\%$ of the entire interaction as context, treating the rest as target.

\paragraph{Conversation Groups as Meta-Learning Tasks?} While our core idea of viewing groups as meta-learning tasks is grounded in social science literature (see Section~\ref{sec:methodology}), does it help to improve empirical performance? Comparing the LL of non-meta-learning and meta-learning models in Table~\ref{tab:nll} by architecture\textemdash VED-MLP against NP and SP-MLP, and VED-GRU against SP-GRU\textemdash we find that accounting for group-specific dynamics through meta-learning yields improved performance. All best-in-family pairwise model differences are statistically significant (Wilcoxon signed rank test, $p < 10^{-4}$).

\paragraph{Comparing Within Meta-Learning Methods.} While our SP-MLP models perform the best on LL in Table~\ref{tab:nll} (pairwise differences are significant), they fare the worst at estimating the mean (Appendix~\ref{app:errors_in_means}). On the other hand, the SP-GRU models estimate a better LL than the NP models with comparable errors in the mean forecast. The NP models attain the lowest errors in predicted means, but also achieve the worst LL. Why do the models achieving better LL also tend to predict worse means? Upon inspecting the metrics for individual features, we found that the models, especially the MLP variants, tend to improve LL by making the variance over constant features exceedingly small, often at the cost of errors in the means. Note that since the rotation in the data is in the horizontal plane, the \textit{qx} and \textit{qy} quaternion dimensions are zero throughout. We do not observe such model behavior in the synthetic data experiments, which do not involve constant features. \autoref{fig:haggling-train-reduced} visualizes forecasts for an example sequence from the Haggling dataset where a turn change has occurred just at the end of the observed window.
Here, the SP-GRU model forecasts an interesting continuation to the turn. It anticipates that the buyer (middle) will interrupt the last observed speaker (right seller), before falling silent and looking from one seller to another, both of whom the model expects to then speak simultaneously (see Appendix~\ref{app:viz} for the full sequence). We believe that the forecast indicates that the model is capable of learning believable haggling turn dynamics from different turn continuations in the data.
From the visualizations also we observe that the models seem to maximize LL at the cost of orientation errors; in the case of SP-MLP seemingly by predicting the majority orientation in the triadic setting. Also, the NP models forecast largely static futures. In contrast, while being more dynamic, the SP-GRU forecasts contain some smoothing. Overall, the SP-GRU models achieve the best trade-off between maximizing LL and forecasting plausible human behavior.

\subsection{Ablations}
\paragraph{Encoding Partner Behavior.}  Modeling the interaction from the perspective of each individual is a central idea in our approach. We investigate the influence of encoding partner behavior into individual representations $\bm{e}^i_\mathrm{ind}$. We train the SP-uniform GRU variant in two configurations: \textit{no-pool}, where we do not encode any partner behavior; and \textit{pool-oT} where we pool over partner representations only at the last timestep (similar to \cite{guptaSocialGANSocially2018}).
Both configurations lead to worse LL and location errors (Table~\ref{tab:ablations-nll} and Appendix~\ref{app:results}).

\begin{table}[!t]
\ra{1.1}
\centering
\scriptsize
\caption{\textbf{Mean (Std.) LL for the Ablation Experiments with the SP-uniform GRU Model.} The reported mean and std. are over individual sequences in the test sets. Higher is better.}
\label{tab:ablations-nll}

\begin{tabular*}{\textwidth}{@{}l @{\extracolsep{\fill}} lcccc@{}}
\toprule
  & & \multicolumn{2}{c}{\textbf{MatchNMingle}} & \multicolumn{2}{c}{\textbf{Haggling}} \\
  \cmidrule{3-6}
  & & \textbf{Random} & \textbf{Fixed-Initial} & \textbf{Random} & \textbf{Fixed-Initial} \\
  
\midrule
  \textbf{Full Model} & & 87.0~(38.4) & 85.5~(38.3) & 79.9~(50.5) & 78.6~(52.2) \\
  
\specialrule{\lightrulewidth}{1ex}{1ex}
  \multirow{2}{*}{\textbf{Encoding Partner Behavior}}
  & no-pool & 77.8~(31.2) & 76.9~(31.0) & 54.5~(75.5) & 50.1~(97.5) \\
  & pool-oT & 82.3~(33.3) & 81.0~(33.6) & 66.9~(26.0) & 66.8~(25.7) \\ [1ex]

  \multirow{2}{*}{\textbf{No Deterministic Decoding}} 
  & Shared Social Encoders   & 88.5~(40.7) & 87.6~(39.6) & 93.1~(39.3) & 91.9~(40.4) \\
  & Unshared Social Encoders & 81.4~(38.1) & 80.2~(37.8) & 66.6~(24.0) & 64.8~(23.4) \\
\bottomrule
\end{tabular*}
\end{table}

\paragraph{Deterministic Decoding and Social Encoder Sharing.}
We investigate the effect of the deterministic decoders by training the SP-uniform GRU model without them. We also investigate sharing a single social encoder between the Process Encoder and Process Decoder in \autoref{fig:architecture}. Removing the decoders only improves log-likelihood if the encoders are shared, and at the cost of head orientation errors (Table~\ref{tab:ablations-nll} and Appendix~\ref{app:results}).

\section{Discussion}
\label{sec:discussion}
The setting of social conversations remains a uniquely challenging frontier for state-of-the-art low-level behavior forecasting. In the recent forecasting challenge involving dyadic interactions, none of the submitted methods could outperform the naive \textit{zero-velocity} baseline \citep[Sec.~5.5]{palmero2022chalearn}. (The baseline propagates the last observed features into the future as if the person remained static.) Why is this? The predominant focus of researchers working on social human-motion prediction has been pedestrian trajectories \cite{rudenko2020human} or actions such as \textit{punching, kicking, gathering, chasing, etc.} \cite{yaoMultipleGranularityGroup2018, adeli2020socially}. In contrast to such activities which involve pronounced movements, the postural adaptation for regulating conversations is far more subtle (also see the discussion in Appendix~\ref{app:review}). At the same time, the social intelligence required to understand the underlying dynamics that drive a conversation is comparatively more sophisticated than for an action such as a kick. We hope that the social-science considerations informing the design of SCF (joint probabilistic forecasting for all members) and the SP models (groups as meta-learning tasks) constitute a meaningful foundation for future research in this space to build upon. Note that for our task formulation, even the performance of our baseline models constitutes new results.

\paragraph{Cross-Discipline Impact and Ethical Considerations.} While our work here is an \textit{upstream} methodological contribution, the focus on human behavior entails ethical considerations for downstream applications. One such application involves assisting social scientists in developing predictive hypotheses for specific behaviors by examining model predictions.
In these cases, such hypotheses must be verified in subsequent controlled experiments. With the continued targeted development of techniques for recording social behavior in the wild \cite{raman2020modular}, evaluating forecasting models in varied interaction settings would also provide further insight. 
Another application involves helping conversational agents achieve smoother interactions.
Here researchers should be careful that the ability to forecast does not result in nefarious manipulation of user behavior.

\hfill \break
\textbf{Acknowledgements.} This research was partially funded by the Netherlands Organization for
Scientific Research (NWO) under the MINGLE project number 639.022.606.
Chirag would like to thank Amelia Villegas-Morcillo for her input and the innumerable
discussions, and Tiffany Matej Hrkalovic for feedback on parts of the manuscript.

{\small
\bibliographystyle{unsrtnat}
\bibliography{references}
}

\appendix
\onecolumn

\begin{center}
\Large
\textbf{Social Processes: Self-Supervised Meta-Learning\\ over Conversational Groups for\\ Forecasting Nonverbal Social Cues} \\
\smallskip Appendices \\
\smallskip
\end{center}

\setcounter{page}{1}

\section{Detailed Results} \label{app:results}

\subsection{Forecasting Glancing Behavior: Quantitative Results} \label{app:toy-quant}
\begin{figure*}[!htb]
  \centering
  \includegraphics[width=0.95\textwidth]{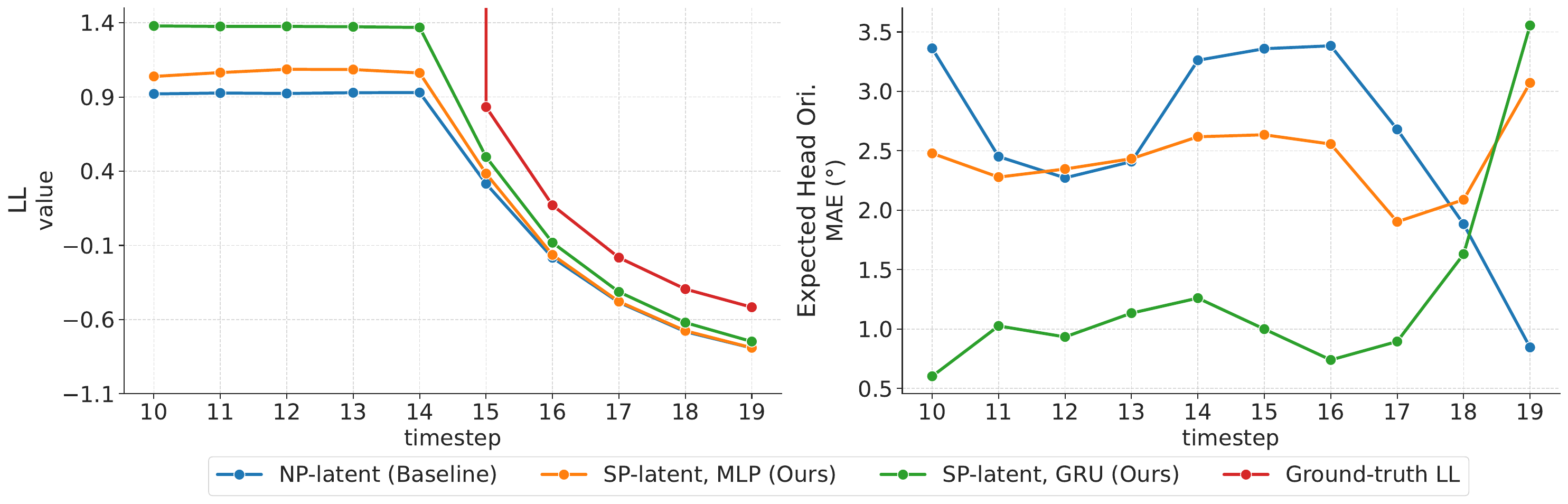}
  \caption{\textbf{Mean Per Timestep Metrics over the Sequences in the Synthetic Glancing Dataset.} We repeat \autoref{fig:ts-synthetic-ll} here for completeness. Head orientation error is computed between the predicted and expected mean (mean of the two ground-truth futures). We observe that the SP-GRU model performs best, especially when the future is certain, learning both the best mean and std. over those timesteps.}
  \label{fig:ts-synthetic}
\end{figure*}

\begin{figure*}[!htb]
  \centering
  \includegraphics[width=\textwidth]{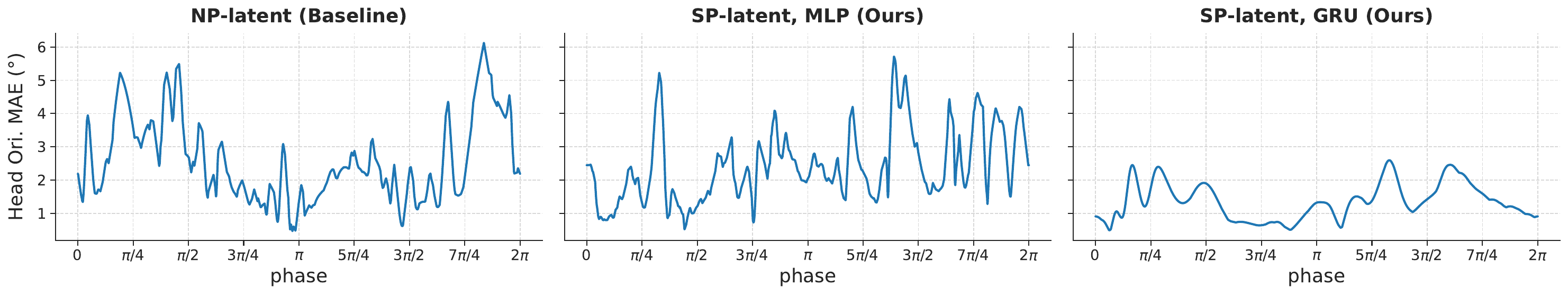}
  \caption{Error in forecast mean and expected mean orientation (mean of the two ground-truth futures) averaged over $\tfut$ for every sequence in the Synthetic Glancing dataset. Each sequence is denoted by the phase of the sinusoid. The SP-GRU error plot is smoother with respect to small phase changes, with lower errors overall.}
  \label{fig:glancing-phases}
\end{figure*}

All models are evaluated under the \textit{random} context regime and \textit{no-pool} configuration. The sinusoids are interpreted to represent a horizontal head rotation between $-90\degree$ and $90\degree$. \autoref{fig:ts-synthetic} plots the LL and head orientation error per timestep in $\tfut$. In \autoref{fig:glancing-phases} we plot the MAE in predicted and expected mean forecasts.

\clearpage
\newpage
\subsection{Errors in Predicted Means for Real-World Behavior Datasets} \label{app:errors_in_means}
Tables~\ref{tab:mean-errors-mnm} and \ref{tab:mean-errors-haggling} specify the error between the predicted mean forecast and ground-truth sequences in the test sets: mean-squared error (MSE) for the head and body keypoint locations; mean absolute error (MAE) in orientation in degrees; and speaking status accuracy. Note that we report the absolute error in rotation in 3D: while the ground-truth normals are constrained to the horizontal plane, we don't constrain our predicted quaternions. The metrics are computed by taking a mean over the participants, timestep dimensions of the tensors. The mean and std. are then reported over individual sequences.

\begin{table*}[!htb]
\ra{1.1}
\centering
\scriptsize
\caption{\textbf{Mean (Std.) Errors in Predicted Means over Sequences in the MatchNMingle Test Sets.} Lower is better. Underline indicates best measure within family.}
\label{tab:mean-errors-mnm}

\begin{subtable}[h]{\textwidth}
\caption{\textbf{Random Context}}
\begin{tabular*}{\textwidth}{@{}l @{\extracolsep{\fill}} lcccc@{}}
\toprule
  \multirow{2}{*}{\textbf{Family}} & \multirow{2}{*}{\textbf{Model}} & \textbf{Head Loc.} & \textbf{Body Loc.} & \textbf{Head Ori.} & \textbf{Body Ori.} \\
  & & MSE (px) & MSE (px) & MAE (\degree) & MAE (\degree) \\

\midrule
  \multirow{2}{*}{\textbf{VED \cite{haNeuralRepresentationSketch2017, bowmanGeneratingSentencesContinuous2016}}}
  & VED-MLP & 131.12~(52.0) & 110.58~(45.3) & 74.84~(39.6) & 97.55~(56.8) \\
  & VED-GRU & \underline{30.00}~(18.0) & \underline{24.80}~(17.2) & \underline{22.62}~(15.7) & \underline{24.72}~(29.1) \\
  
\specialrule{\lightrulewidth}{1ex}{1ex}
  \multirow{4}{*}{\textbf{NP \cite{garneloNeuralProcesses2018, kimAttentiveNeuralProcesses2019}}}
  & NP-latent & \underline{41.16}~(21.4) & \underline{33.41}~(21.1) & 25.58~(21.1) & 35.88~(47.7) \\
  & NP-uniform & 42.93~(20.9) & 39.76~(21.0) & 27.14~(21.4) & 38.22~(48.0) \\
  & ANP-dot & 41.59~(20.4) & 37.57~(19.5) & 26.42~(20.6) & 37.19~(47.6) \\
  & ANP-mh & 41.49~(20.7) & 36.77~(19.5) & \underline{25.39}~(20.7) & \underline{35.82}~(48.0) \\

\specialrule{\lightrulewidth}{1ex}{1ex}
  \multirow{4}{*}{\textbf{Ours (SP-MLP)}}
  & SP-latent & 297.24~(92.3) & 258.71~(87.8) & 95.11~(41.3) & 110.74~(51.6) \\
  & SP-uniform & 73.53~(34.2) & 61.95~(36.2) & \underline{95.00}~(41.3) & 110.06~(50.8) \\
  & ASP-dot & 78.86~(24.8) & 67.77~(21.7) & 95.02~(41.3) & 110.27~(51.1) \\
  & ASP-mh & \underline{63.99}~(22.4) & \underline{53.59}~(22.2) & \underline{95.00}~(41.3) & \underline{109.81}~(50.6) \\ [1ex]
  
  \multirow{4}{*}{\textbf{Ours (SP-GRU)}}
  & SP-latent & \underline{38.58}~(20.9) & \underline{27.45}~(19.9) & \underline{49.50}~(44.1) & 63.06~(57.4) \\
  & SP-uniform & 41.48~(22.2) & 37.82~(18.9) & 56.39~(47.8) & 62.59~(54.3) \\
  & ASP-dot & 44.17~(21.4) & 37.13~(19.7) & 55.41~(47.1) & 62.14~(54.5) \\
  & ASP-mh & 43.49~(21.3) & 38.29~(19.8) & 57.68~(47.3) & \underline{61.94}~(53.7) \\
\bottomrule
\end{tabular*}
\end{subtable}

\vfill
\begin{subtable}[h]{\textwidth}
\caption{\textbf{Fixed-Initial Context}}
\begin{tabular*}{\textwidth}{@{}l @{\extracolsep{\fill}} lcccc@{}}
\toprule
  \multirow{2}{*}{\textbf{Family}} & \multirow{2}{*}{\textbf{Model}} & \textbf{Head Loc.} & \textbf{Body Loc.} & \textbf{Head Ori.} & \textbf{Body Ori.} \\
  & & MSE (px) & MSE (px) & MAE (\degree) & MAE (\degree) \\
  
\midrule
  \multirow{2}{*}{\textbf{VED \cite{haNeuralRepresentationSketch2017, bowmanGeneratingSentencesContinuous2016}}}
  & VED-MLP & 131.67~(52.6) & 111.50~(46.0) & 75.97~(38.7) & 98.26~(55.6) \\
  & VED-GRU & \underline{29.51}~(16.5) & \underline{24.33}~(16.1) & \underline{22.75}~(15.9) & \underline{26.60}~(32.0) \\

\specialrule{\lightrulewidth}{1ex}{1ex}
  \multirow{4}{*}{\textbf{NP \cite{garneloNeuralProcesses2018, kimAttentiveNeuralProcesses2019}}}
  & NP-latent &  \underline{40.82}~(19.1) & \underline{32.81}~(19.5) & \underline{25.58}~(21.6) & \underline{38.97}~(51.0) \\
  & NP-uniform & 45.22~(19.1) & 40.60~(19.0) & 28.34~(22.4) & 42.65~(51.8) \\
  & ANP-dot & 44.67~(18.6) & 40.03~(18.2) & 29.08~(21.9) & 44.44~(54.4) \\
  & ANP-mh & 42.75~(18.7) & 37.56~(18.4) & 26.95~(22.3) & 42.20~(51.9) \\

\specialrule{\lightrulewidth}{1ex}{1ex}
  \multirow{4}{*}{\textbf{Ours (SP-MLP)}}
  & SP-latent & 296.36~(92.8) & 259.46~(87.5) & 94.75~(39.0) & 108.62~(47.3) \\
  & SP-uniform & 81.61~(40.7) & 64.44~(42.6) & 94.68~(39.0) & 108.26~(46.7) \\
  & ASP-dot & 92.03~(38.2) & 78.97~(33.1) & 94.69~(39.0) & 108.36~(46.9) \\
  & ASP-mh & \underline{66.22}~(25.5) & \underline{53.04}~(24.0) & \underline{94.67}~(39.0) & \underline{108.14}~(46.5) \\ [1ex]
  
  \multirow{4}{*}{\textbf{Ours (SP-GRU)}}
  & SP-latent & \underline{38.31}~(18.2) & \underline{26.79}~(17.7) & \underline{51.78}~(45.1) & 65.38~(55.9) \\
  & SP-uniform & 42.75~(21.7) & 42.18~(19.8) & 57.79~(48.6) & \underline{64.44}~(53.3) \\
  & ASP-dot & 54.42~(25.9) & 44.88~(22.6) & 56.12~(46.9) & 65.28~(54.5) \\
  & ASP-mh & 56.62~(26.3) & 47.78~(22.9) & 58.90~(47.9) & 64.46~(54.1) \\
\bottomrule
\end{tabular*}
\end{subtable}
\end{table*}

The keypoint annotations for MnM are provided in image space from a top-down perspective, so the location errors in Table~\ref{tab:mean-errors-mnm} are reported as the MSE in pixel locations. 
We do not consider speaking status cues for experiments with MnM (see Appendix~\ref{app:preprocessing_realworld}). 

\begin{table*}[!htb]
\ra{1.1}
\centering
\scriptsize
\caption{\textbf{Mean (Std.) Errors in Predicted Means over Sequences in the Haggling Test Sets.} Lower is better for all metrics except for speaking status accuracy. Underline indicates best measure within family.}
\label{tab:mean-errors-haggling}

\begin{subtable}[h]{\textwidth}
\caption{\textbf{Random Context}}
\begin{tabular*}{\textwidth}{@{}l @{\extracolsep{\fill}} lccccc@{}}
\toprule
  \multirow{2}{*}{\textbf{Family}} & \multirow{2}{*}{\textbf{Model}} & \textbf{Head Loc.} & \textbf{Body Loc.} & \textbf{Head Ori.} & \textbf{Body Ori.} & \textbf{Speaking} \\
  & & MSE (cm) & MSE (cm) & MAE (\degree) & MAE (\degree) & Accuracy \\

\midrule
  \multirow{2}{*}{\textbf{VED \cite{haNeuralRepresentationSketch2017, bowmanGeneratingSentencesContinuous2016}}}
  & VED-MLP & 42.04~(16.0) & 41.53~(15.6) & 24.70~(20.7) & 19.02~(13.3) & 0.636~(0.24) \\
  & VED-GRU & \underline{0.79}~(0.4) & \underline{0.75}~(0.4) & \underline{1.55}~(0.6) & \underline{1.06}~(0.4) & \underline{0.989}~(0.02) \\
  
\specialrule{\lightrulewidth}{1ex}{1ex}
  \multirow{4}{*}{\textbf{NP \cite{garneloNeuralProcesses2018, kimAttentiveNeuralProcesses2019}}}
  & NP-latent & 14.21~(6.5) & 15.06~(6.1) & 16.29~(13.8) & 12.82~(13.7) & 0.787~(0.23) \\
  & NP-uniform & 15.01~(7.3) & 15.97~(7.2) & 17.45~(18.3) & 14.65~(20.0) & 0.715~(0.24) \\
  & ANP-dot & \underline{11.86}~(5.4) & \underline{12.22}~(5.5) & \underline{15.44}~(13.3) & \underline{12.56}~(18.0) & \underline{0.806}~(0.23) \\
  & ANP-mh & 16.36~(7.4) & 17.17~(7.2) & 19.41~(20.4) & 16.02~(22.1) & 0.692~(0.21) \\

\specialrule{\lightrulewidth}{1ex}{1ex}
  \multirow{4}{*}{\textbf{Ours (SP-MLP)}}
  & SP-latent & 25.58~(10.1) & \underline{26.57}~(9.0) & 91.07~(23.9) & 97.09~(22.5) & 0.638~(0.08) \\
  & SP-uniform & 31.99~(8.2) & 36.33~(7.3) & 91.08~(23.9) & 91.36~(23.9) & 0.629~(0.18) \\
  & ASP-dot & 27.16~(7.7) & 31.19~(7.1) & 90.88~(23.9) & 91.43~(23.8) & 0.704~(0.19) \\
  & ASP-mh & \underline{23.88}~(7.8) & 27.13~(7.7) & \underline{90.50}~(23.9) & \underline{91.04}~(24.1) & \underline{0.792}~(0.24) \\ [1ex]
  
  \multirow{4}{*}{\textbf{Ours (SP-GRU)}}
  & SP-latent & 17.18~(6.5) & 17.41~(6.2) & \underline{17.76}~(15.8) & \underline{14.78}~(20.7) & 0.713~(0.23) \\
  & SP-uniform & 15.84~(5.5) & 17.76~(7.5) & 20.65~(19.9) & 21.73~(29.5) & 0.671~(0.22) \\
  & ASP-dot & 22.59~(8.7) & 23.52~(10.2) & 17.90~(11.3) & 16.10~(19.3) & 0.722~(0.24) \\
  & ASP-mh & \underline{14.65}~(5.8) & \underline{15.38}~(6.1) & 28.06~(24.5) & 36.90~(37.9) & \underline{0.767}~(0.23) \\
\bottomrule
\end{tabular*}
\end{subtable}

\vfill
\begin{subtable}[h]{\textwidth}
\caption{\textbf{Fixed-Initial Context}}
\begin{tabular*}{\textwidth}{@{}l @{\extracolsep{\fill}} lccccc@{}}
\toprule
  \multirow{2}{*}{\textbf{Family}} & \multirow{2}{*}{\textbf{Model}} & \textbf{Head Loc.} & \textbf{Body Loc.} & \textbf{Head Ori.} & \textbf{Body Ori.} & \textbf{Speaking} \\
  & & MSE (cm) & MSE (cm) & MAE (\degree) & MAE (\degree) & Accuracy \\

\midrule
  \multirow{2}{*}{\textbf{VED \cite{haNeuralRepresentationSketch2017, bowmanGeneratingSentencesContinuous2016}}}
  & VED-MLP & 41.71~(16.2) & 41.27~(15.8) & 24.36~(19.8) & 19.33~(13.4) & 0.640~(0.25) \\
  & VED-GRU & \underline{0.76}~(0.4) & \underline{0.72}~(0.3) & \underline{1.56}~(0.6) & \underline{1.04}~(0.3) & \underline{0.989}~(0.02) \\

\specialrule{\lightrulewidth}{1ex}{1ex}
  \multirow{4}{*}{\textbf{NP \cite{garneloNeuralProcesses2018, kimAttentiveNeuralProcesses2019}}}
  & NP-latent & 13.85~(6.1) & 14.71~(5.7) & 16.22~(14.1) & \underline{12.69}~(13.9) & \underline{0.774}~(0.24) \\
  & NP-uniform & 15.01~(7.5) & 15.95~(7.5) & 17.26~(15.9) & 14.68~(18.7) & 0.701~(0.24) \\
  & ANP-dot & \underline{12.83}~(5.9) & \underline{13.26}~(6.0) & \underline{16.19}~(13.7) & 13.56~(17.8) & 0.717~(0.23) \\
  & ANP-mh & 16.68~(7.9) & 17.43~(7.7) & 19.78~(21.2) & 15.57~(20.3) & 0.682~(0.21) \\

\specialrule{\lightrulewidth}{1ex}{1ex}
  \multirow{4}{*}{\textbf{Ours (SP-MLP)}}
  & SP-latent & 25.27~(10.0) & \underline{26.33}~(8.9) & 91.14~(23.8) & 97.09~(22.5) & 0.640~(0.09) \\
  & SP-uniform & 32.93~(9.4) & 37.16~(8.5) & 91.15~(23.9) & 91.36~(23.9) & 0.633~(0.18) \\
  & ASP-dot & 27.94~(7.8) & 31.83~(7.1) & 90.93~(23.9) & 91.43~(23.8) & 0.628~(0.20) \\
  & ASP-mh & \underline{24.07}~(8.1) & 27.35~(8.3) & \underline{90.53}~(23.9) & \underline{91.07}~(24.1) & \underline{0.770}~(0.25) \\ [1ex]
  
  \multirow{4}{*}{\textbf{Ours (SP-GRU)}}
  & SP-latent & 16.66~(6.2) & 17.17~(6.0) & \underline{17.67}~(16.0) & \underline{14.64}~(20.3) & \underline{0.705}~(0.23)   \\
  & SP-uniform & \underline{16.53}~(6.0) & 18.20~(8.0) & 20.74~(19.5) & 21.31~(28.9) & 0.674~(0.22)   \\
  & ASP-dot & 23.91~(8.8) & 25.34~(10.6) & 19.11~(12.8) & 17.36~(19.0) & 0.635~(0.26)  \\
  & ASP-mh & 16.87~(6.0) & \underline{16.96}~(6.1) & 28.90~(24.3) & 37.23~(37.6) & \underline{0.705}~(0.24) \\
\bottomrule
\end{tabular*}
\end{subtable}
\end{table*}

\clearpage
\newpage
\subsection{Ablations} \label{app:ablations}

\begin{table*}[!htb]
\vspace{-30pt}
\ra{1.1}
\centering
\scriptsize
\caption{\textbf{Mean (Std.) Errors in Predicted Means for the Ablation Experiments with the SP-uniform GRU Model.} The reported mean and std. are over sequences in the MatchNMingle Test Sets. Lower is better.}
\label{tab:ablations-means-mnm}

\begin{subtable}[t]{\textwidth}
\caption{\textbf{Random Context}}
\begin{tabular*}{\textwidth}{@{}l @{\extracolsep{\fill}} lcccc@{}}
\toprule
  & & \textbf{Head Loc.} & \textbf{Body Loc.} & \textbf{Head Ori.} & \textbf{Body Ori.} \\
  & & MSE (px) & MSE (px) & MAE (\degree) & MAE (\degree) \\

\midrule
  \textbf{Full Model} & & 41.48~(22.2) & 37.82~(18.9) & 56.39~(47.8) & 62.59~(54.3) \\
   
\specialrule{\lightrulewidth}{1ex}{1ex}
  \multirow{2}{*}{\textbf{Encoding Partner Behavior}}
  & no-pool & 36.25~(19.3) & 30.88~(18.1) & 47.28~(39.0) & 62.09~(54.5) \\
  & pool-oT & 41.81~(19.7) & 33.78~(17.9) & 54.01~(45.3) & 63.32~(54.8) \\ [1ex]

  \multirow{2}{*}{\textbf{No Deterministic Decoding}}
  & Shared Social Encoders   & 41.84~(19.9) & 30.99~(18.4) & 44.59~(37.2) & 72.02~(62.4) \\
  & Unshared Social Encoders & 37.25~(19.7) & 36.13~(18.0) & 62.81~(55.6) & 56.15~(52.6) \\
\bottomrule
\end{tabular*}
\end{subtable}

\begin{subtable}[t]{\textwidth}
\caption{\textbf{Fixed-Initial Context}}
\begin{tabular*}{\textwidth}{@{}l @{\extracolsep{\fill}} lcccc@{}}
\toprule
  & & \textbf{Head Loc.} & \textbf{Body Loc.} & \textbf{Head Ori.} & \textbf{Body Ori.} \\
  & & MSE (px) & MSE (px) & MAE (\degree) & MAE (\degree) \\

\midrule
   \textbf{Full Model} & & 42.75~(21.7) & 42.18~(19.8) & 57.79~(48.6) & 64.44~(53.3) \\
   
\specialrule{\lightrulewidth}{1ex}{1ex}
  \multirow{2}{*}{\textbf{Encoding Partner Behavior}}
  & no-pool & 36.17~(17.4) & 31.77~(16.5) & 48.28~(39.6) & 64.19~(53.3) \\
  & pool-oT & 41.91~(18.6) & 34.17~(16.0) & 54.95~(45.7) & 65.20~(53.7) \\ [1ex]

  \multirow{2}{*}{\textbf{No Deterministic Decoding}}
  & Shared Social Encoders   & 41.29~(18.2) & 31.62~(16.9) & 45.54~(38.0) & 73.30~(60.9) \\
  & Unshared Social Encoders & 37.78~(18.5) & 35.28~(16.3) & 63.96~(56.1) & 58.23~(53.0) \\
\bottomrule
\end{tabular*}
\end{subtable}
\end{table*}

\begin{table*}[!htb]
\vspace{-30pt}
\ra{1.1}
\centering
\scriptsize
\caption{\textbf{Mean (Std.) Errors in Predicted Means for the Ablation Experiments with the SP-uniform GRU Model.} The reported mean and std. are over sequences in the Haggling Test Sets. Lower is better for all except for speaking status accuracy.}
\label{tab:ablations-means-haggling}

\begin{subtable}[t]{\textwidth}
\caption{\textbf{Random Context}}
\begin{tabular*}{\textwidth}{@{}l @{\extracolsep{\fill}} lccccc@{}}
\toprule
  & & \textbf{Head Loc.} & \textbf{Body Loc.} & \textbf{Head Ori.} & \textbf{Body Ori.} & \textbf{Speaking} \\
  & & MSE (cm) & MSE (cm) & MAE (\degree) & MAE (\degree) & Accuracy \\

\midrule
   \textbf{Full Model} & & 15.84~(5.5) & 17.76~(7.5) & 20.65~(19.9) & 21.73~(29.5) & 0.671~(0.22) \\
   
\specialrule{\lightrulewidth}{1ex}{1ex}
  \multirow{2}{*}{\textbf{Encoding Partner Behavior}}
  & no-pool & 18.20~(6.7) & 18.05~(7.7) & 16.76~(12.8) & 14.30~(20.9) & 0.690~(0.21) \\
  & pool-oT & 17.02~(6.1) & 19.18~(6.5) & 23.71~(25.1) & 17.80~(26.8) & 0.738~(0.21) \\ [1ex]

  \multirow{2}{*}{\textbf{No Deterministic Decoding}}
  & Shared Social Encoders   & 15.76~(7.2) & 16.34~(6.6) & 45.54~(44.6) & 21.87~(25.0) & 0.644~(0.22) \\
  & Unshared Social Encoders & 17.40~(6.9) & 18.33~(6.7) & 18.62~(14.7) & 14.54~(20.2) & 0.704~(0.23) \\
\bottomrule
\end{tabular*}
\end{subtable}

\begin{subtable}[t]{\textwidth}
\caption{\textbf{Fixed-Initial Context}}
\begin{tabular*}{\textwidth}{@{}l @{\extracolsep{\fill}} lccccc@{}}
\toprule
  & & \textbf{Head Loc.} & \textbf{Body Loc.} & \textbf{Head Ori.} & \textbf{Body Ori.} & \textbf{Speaking} \\
  & & MSE (cm) & MSE (cm) & MAE (\degree) & MAE (\degree) & Accuracy \\

\midrule
   \textbf{Full Model} & & 16.53~(6.0) & 18.20~(8.0) & 20.74~(19.5) & 21.31~(28.9) & 0.674~(0.22) \\
   
\specialrule{\lightrulewidth}{1ex}{1ex}
  \multirow{2}{*}{\textbf{Encoding Partner Behavior}}
  & no-pool & 18.64~(6.7) & 18.45~(7.4) & 16.85~(12.9) & 14.29~(20.5) & 0.687~(0.21) \\
  & pool-oT & 17.39~(6.2) & 18.97~(6.4) & 23.90~(24.6) & 17.63~(25.6) & 0.730~(0.21) \\ [1ex]

  \multirow{2}{*}{\textbf{No Deterministic Decoding}}
  & Shared Social Encoders   & 16.93~(8.1) & 17.15~(7.0) & 45.49~(44.3) & 21.83~(24.7) & 0.637~(0.22) \\
  & Unshared Social Encoders & 18.54~(7.9) & 19.18~(7.1) & 18.68~(14.9) & 14.44~(20.0) & 0.700~(0.23) \\
\bottomrule
\end{tabular*}
\end{subtable}
\end{table*}

\clearpage
\section{Qualitative Visualizations} \label{app:viz}

\subsection{Glancing Behavior}

\begin{figure*}[!h]
  \centering
  \includegraphics[width=\textwidth]{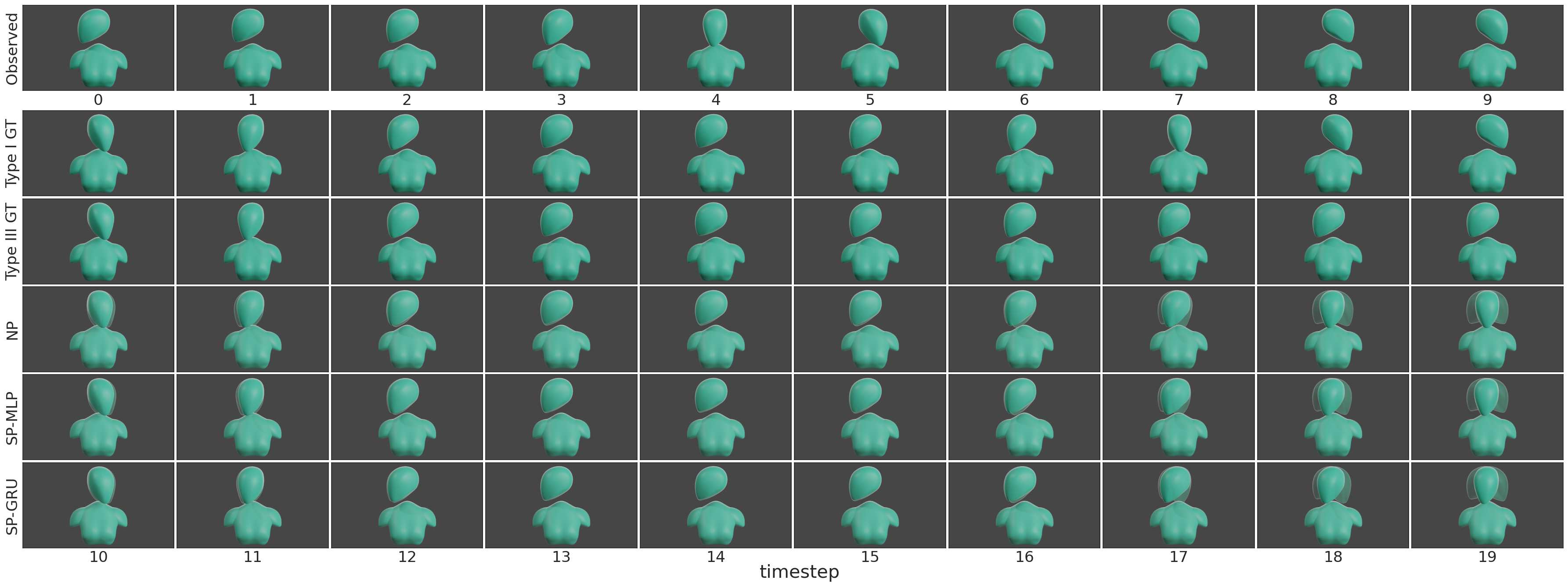}
  \caption{\textbf{Forecasting Glancing Behavior for a Sequence in the Context Set.} We visualize the same sinusoid within the context set as plotted in \autoref{fig:toy} ($\text{phase}=4.2$), here interpreted as a horizontal head rotation between $-90\degree$ and $90\degree$. The bottom three rows depict predictions, with the solid head denoting the mean, and the translucent heads the std. \textit{GT} stands for \textit{Ground-Truth}. The SP models learn better uncertainty estimates, especially over the timesteps where the future is certain (see timestep 11, for instance).}
  \label{fig:qual-glance-tc}
\end{figure*}

\begin{figure*}[!h]
  \centering
  \includegraphics[width=\textwidth]{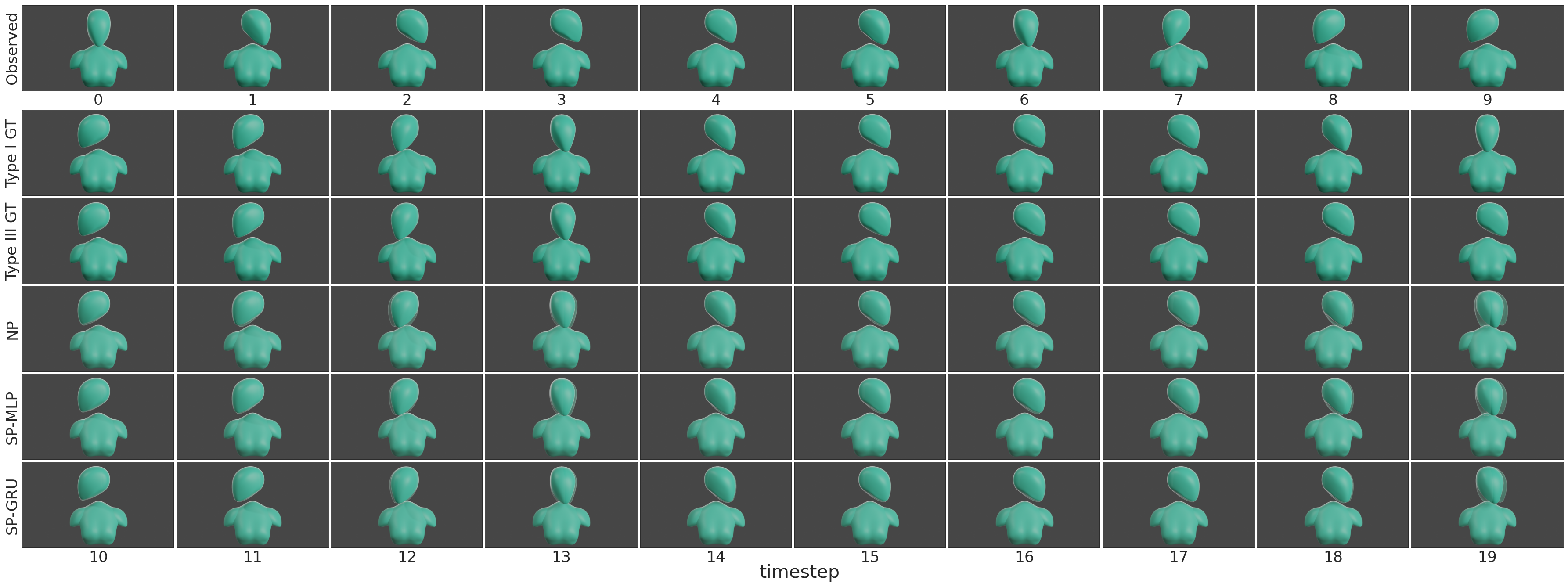}
  \caption{\textbf{Forecasting Glancing Behavior for a Sequence Not in the Context Set.} We visualize the same sinusoid not in the context set as plotted in \autoref{fig:toy} ($\text{phase}=0.005$). See the \autoref{fig:qual-glance-tc} caption for details.}
  \label{fig:qual-glance-tnc}
\end{figure*}

\clearpage
\subsection{Real-World Behavior}

\begin{figure*}[!h]
  \centering
  \includegraphics[width=\textwidth]{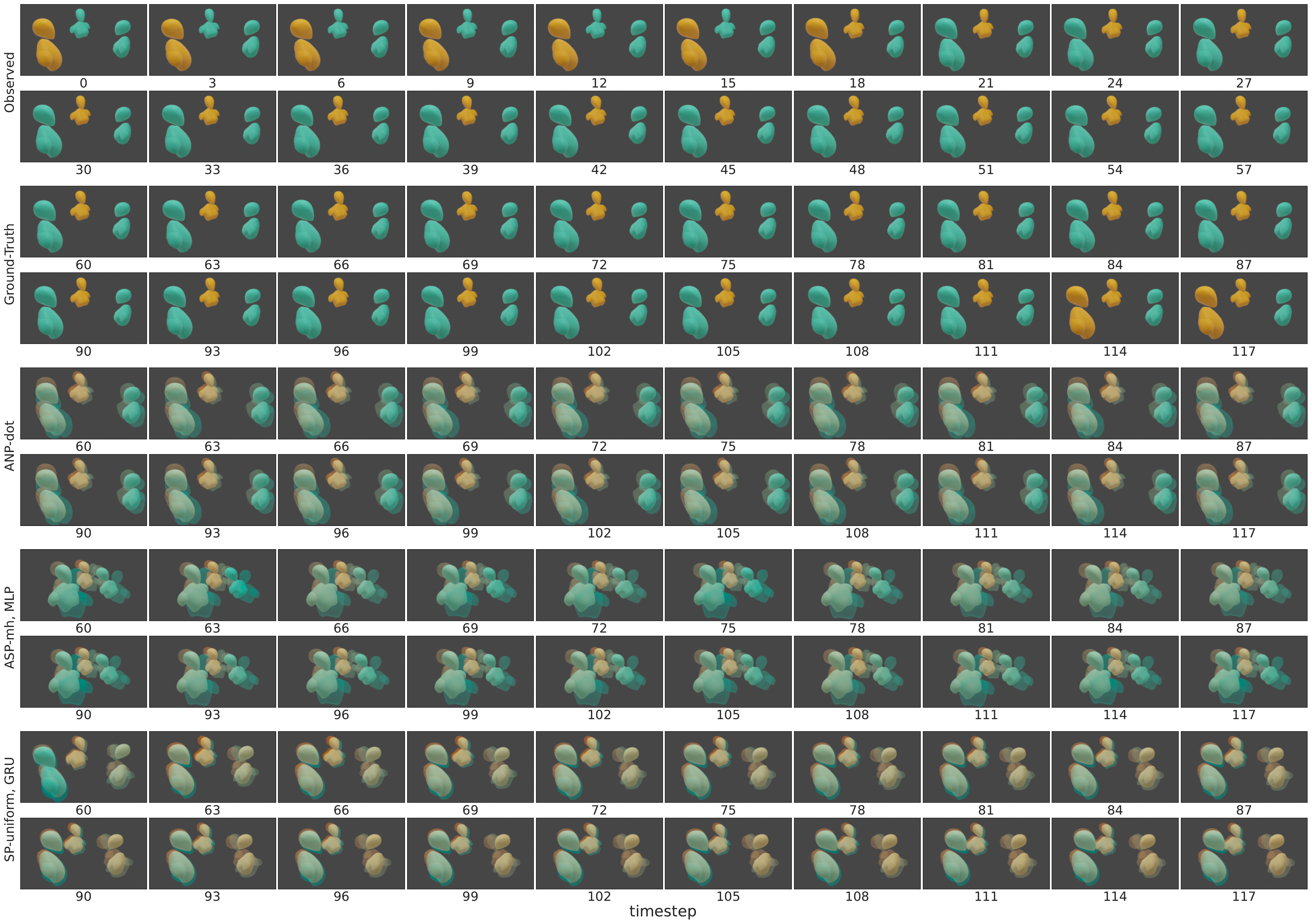}
  \caption{\textbf{Forecasts for a Sequence from the Haggling Test Group \textit{170221-b1-group3}.} We visualize cues from real-world data using 3D models to preserve privacy. Similar to \autoref{fig:haggling-train-reduced}, speakers are depicted in orange. The predicted speaking status mean is visualized as an interpolated shade between orange and blue. The translucent models in the forecasts denote the $\text{mean} \pm \text{std.}$ pose and speaking status. A speaker turn change occurs at around timestep $18$ in the observed sequence. The buyer (on the right) looks at both sellers in turn mostly through gaze changes visible in the original video. This is barely registered in posture changes since both speakers are within the buyer's field of vision in this triadic setting (see Appendix~\ref{app:review} for a discussion). The leaning motions of the new speaker, however, are captured in the postural shifts that continue into the ground-truth future. We observe that the NP forecasts are almost completely static. The SP-GRU forecasts are comparatively dynamic with lower uncertainties overall. The SP-MLP model seems to be learning an overall average orientation, forecasting all participants to be facing in the direction of the two sellers. Note that these pose changes are far more subtle than in the glancing behavior dataset, which is an important consideration for the domain practice of evaluating methods on synthesized behavior alone \cite{vazquezMaintainingAwarenessFocus2016, sanghvi2019mgpi}.}
  \label{fig:haggling-test}
\end{figure*}

\clearpage
\begin{figure*}[!h]
  \centering
  \includegraphics[width=\textwidth]{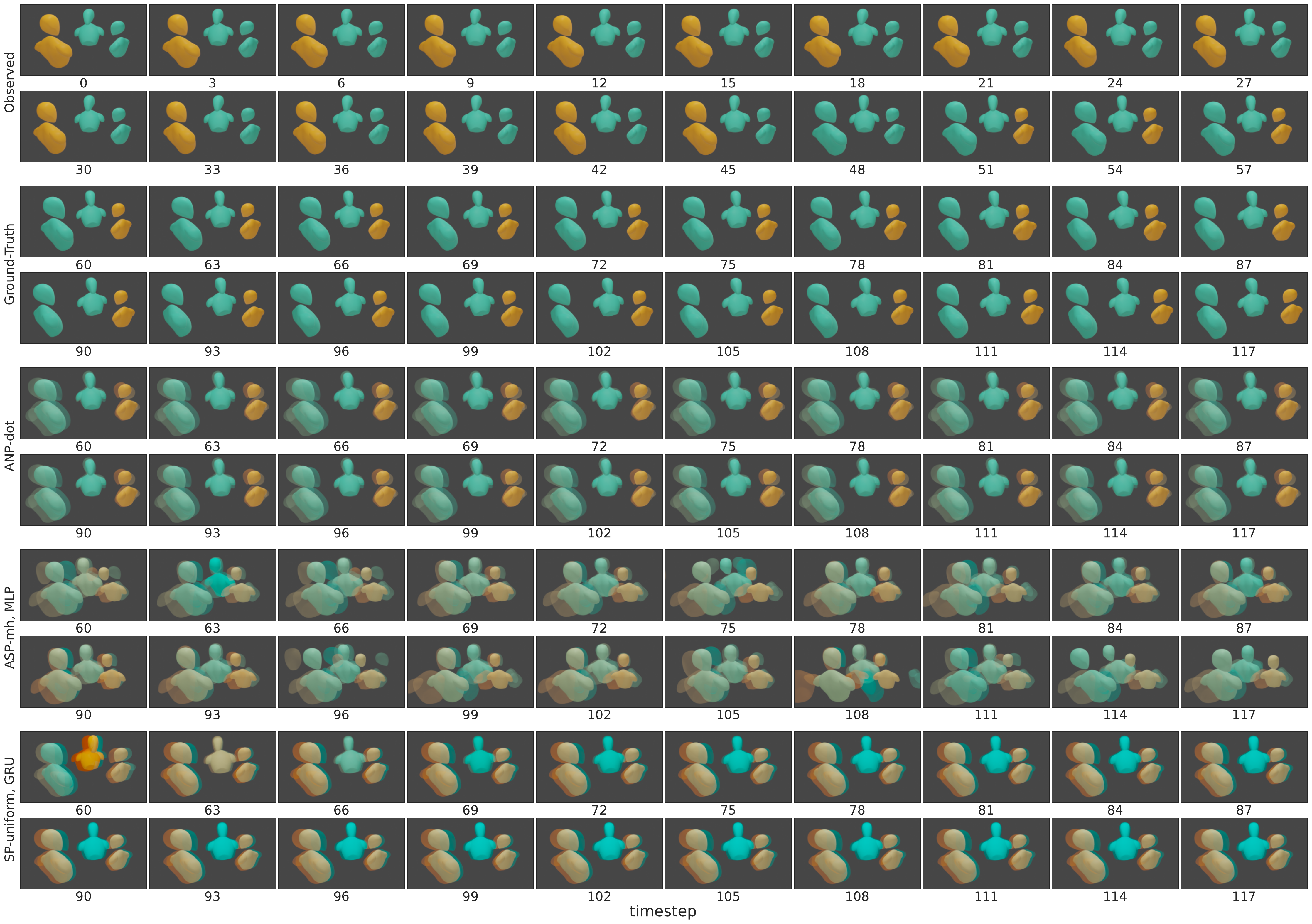}
  \caption{\textbf{Forecasts for a Sequence from the Haggling Train Group \textit{170224-a1-group1}.} We see a similar pattern to the model forecasts as in \autoref{fig:haggling-test}: NP forecasts are static, SP-GRU predicts more dynamic futures, while the SP-MLP forecasts average orientations. A turn change has occurred at the end of the observed window. We observe that the SP-GRU model forecasts an interesting continuation of the turn. It anticipates the buyer (middle) to quickly interject the last observed speaking seller, before falling silent and directing attention between the sellers, both of whom it expects to then speak simultaneously. While this is not the ground-truth future in this instance, we believe that the forecast still indicates that the model is capable of learning believable haggling turn dynamics from the overall training data. See the \autoref{fig:haggling-test} for details on the visualization setup.}
  \label{fig:haggling-train}
\end{figure*}

\clearpage
\newpage
\section{Additional Dataset Details} \label{app:data}

\subsection{Synthesized Glancing Behavior Dataset} \label{app:toy-set}
The set of pristine sinusoids representing \textit{Type I} glances is computed by evaluating the sine function at the bounds of $19$ equally spaced partitions of $[0, 3\pi + \phi)$, for phase values $\phi$ in $[0, 2\pi)$ with a step size of $0.001$. More concretely, this is the set
\begin{multline}
    g = \{ r: r = \sin(x),\ x = n \times (3\pi + \phi) / 19,\ n \in \{0, 1, \ldots 19\},\
    \\
    \phi = p \times 0.001,\ p \in \{0, 1, \dots 6283\}\},
\end{multline}
which results in $6284$ sequences. \textit{Type III} glances are represented by identical sinusoids with clipped amplitudes for the last six timesteps, resulting in the final dataset of $12568$ sequences. We train with batches of $100$ sequences, using a randomly sampled $25~\%$ of the batch as context. For evaluation, we fix $785$ randomly sampled phase values as context. For each phase, samples corresponding to both types of glances are included in the context set, effectively using $25~\%$ of all samples as context at evaluation.   

\subsection{Preprocessing the Real-World Behavior Datasets} \label{app:preprocessing_realworld}
For MnM, 2D keypoints are provided in image space (from a top-down camera perspective). For Haggling, the keypoints are in 3D space, and we use the nose keypoint to represent the head location, and the mid-point of the shoulders to represent the body location.
We standardize the location features to have zero mean and unit variance, using the train statistics to standardize the test sets.

\paragraph{Haggling Preprocessing Details.} Cue annotations are provided at $30$~Hz for the Haggling Dataset.
Motivated by the domain focus on the organization of turn-taking, we consider window lengths of $2$ seconds supported by dataset statistics and literature. The Haggling dataset duration of contiguous speech follows a mean of $2.13$~s ($\sigma = 2.61$~s), which is close to the mean measure of $1.68$~s found in turn-taking analysis \cite{heldnerPausesGapsOverlaps2010, raman2019towards}. We generate sliding windows with an overlap of $0.8$, constraining the offset between $\tobs$ and $\tfut$ to a maximum of $5$~s. This is to roughly restrict candidate future windows to those starting after two turn changes. In total, we obtain about $135$K observed-future sequence pairs for training, and about $48$K pairs for testing.

\paragraph{MatchNMingle Preprocessing Details.} Cue annotations in MnM are provided at $1$~Hz. The provided speaking status labels were annotated from video alone, and then manually smoothed by majority voting over $3$~s windows. Consequently, these often do not match a person's pose behavior in the video for long sequences. We therefore deemed this data stream unsuitable for continuous sequence prediction and excluded it from our experiments. Assuming about $2$~s per turn as before, and considering the $1$~Hz annotation sample rate, we choose $\tobs$ and $\tfut$ to contain $4$ timesteps or two turn durations each, with a maximum offset of $4$~s as well. The keypoint annotations for every person are provided within the camera that best captures the individual, which can change over the duration of the interaction. For every group, we therefore first extract slices where the entire group is visible within the same camera for at least $20$~s. We found $20$~s to be a reasonable balance between not aggressively discarding groups, while still obtaining unique observed sequences for each slice (at least four). In total, we obtained about $74$K observed-future sequence pairs for training, and about $52$K pairs for testing.

\section{Implementation Details} \label{app:implementation}

\subsection{Neural Architectures}
The hyperparameters we chose resulted from light tuning through $5$-fold cross-validation and showed improved performance for all models, but improved absolute performance might be obtained through more extensive tuning. The architecture hyperparameters were then kept fixed for the variants within each family for fair intra-family comparison. Table~\ref{tab:hparams} specifies the network architecture hyperparameters for the real-world behavior dataset experiments. Note that for the MLP variants, the number of parameters is dependent on sequence length (timestep and feature dimensions of the tensors are collapsed into a single dimension; $60$ timesteps for Haggling, $4$ for MnM), so the final number of parameters vary across the datasets.

\begin{table}[!htb]
\ra{1.1}
\centering
\scriptsize
\caption{\textbf{Architecture Hyperparameters for real-world behavior dataset experiments (MnM / Haggling).} For the meta-learning models, the number of parameters are reported for the simplest \textit{-latent} variant.}
\label{tab:hparams}

\begin{tabular}[b]{@{}lccccc@{}}
\toprule
  \textbf{Hyperparameter} & \textbf{VED-MLP} & \textbf{VED-GRU} & \textbf{NP} & \textbf{SP-MLP} & \textbf{SP-GRU} \\

\midrule
  \textbf{Sequence Encoder/Decoder} \\
    \hspace{3mm}Number of layers        & $2$ & $1$ & $2$ & $2$ & $1$ \\
    \hspace{3mm}Hidden dim              & $180$ & $320$ & $180/460$ & $64$ & $320$ \\
  \textbf{Partner Pooler $\psi(\bm{x}_j)$} \\
    \hspace{3mm}Number of MLP layers    & \textemdash & \textemdash & \textemdash & $2$ & $2$ \\
    \hspace{3mm}MLP hidden dim          & \textemdash & \textemdash & \textemdash & $64$ & $64$ \\
    \hspace{3mm}Output dim              & \textemdash & \textemdash & \textemdash & $32$ & $32$ \\
  \textbf{$\bm{z}$ Encoder} \\
    \hspace{3mm}Number of layers        & $2$ & $2$ & $2$ & $2$ & $2$ \\
    \hspace{3mm}Hidden dim              & $64$ & $64$ & $64$ & $64$ & $64$ \\
  \textbf{Representations} \\
    \hspace{3mm}$\bm{e}$, $\bm{r}$, $\bm{s}$, $\bm{z}$ dim  & $64$ & $64$ & $64$ & $64$ & $64$ \\
  \textbf{Multi-Head Attention} \\
    \hspace{3mm}Query/Key dim           & \textemdash & \textemdash & $32$ & $32$ & $32$ \\
    \hspace{3mm}Number of heads         & \textemdash & \textemdash & $8$ & $8$ & $8$ \\

\midrule
  \textbf{Number of parameters} \\
  \hspace{3mm}MatchNMingle Dataset &  $254$K & $1.1$M & $274$K & $283$K & $3.0$M \\
  \hspace{3mm}Haggling Dataset &  $711$K & $1.1$M & $2.8$M & $2.2$M & $3.0$M \\
\bottomrule
\end{tabular}
\end{table}

The non-meta-learning baselines retain the probabilistic attributes of our proposed Social Process models so that the only difference is the meta-learning aspect. We consequently adapt these baseline models from RNN based variational autoencoder architectures, first proposed for autoencoding sentences \cite{bowmanGeneratingSentencesContinuous2016}, and later refined for sketches \cite{haNeuralRepresentationSketch2017}. The key difference is that rather than autoencoding the observed cues, we decode the future cues from the latent representations. Unlike \cite{bowmanGeneratingSentencesContinuous2016}, we are not working with discrete inputs, so the cues are fed directly into the sequence encoders without an embedding layer. For consistent comparison across models, we use unidirectional sequence encoders and decoders for the GRU variants and omit the Gaussian Mixture Model layer of \cite{haNeuralRepresentationSketch2017}. This way, the encoding of partner behavior is the only architectural difference in the backbone components between our proposed SP models and the VED baselines.

\subsection{Training and Evaluation} \label{app:training}
The models are trained in the \textit{random} context regime following the standard NP setting.
We construct batches for training by bucketing samples such that all sequences in a batch share the same length of $\tobs$ and $\tfut$. Note that since the MLP models are operationalized by collapsing the timestep and feature dimensions, the length of $\tfut$ is fixed for these models across batches. However, since the recurrent models can handle sequences of different lengths, we allow for forecasting different length futures across batches, resulting in a few more training batches. Following the training practices suggested by \citet{leEmpiricalEvaluationNeural}, we construct the context set at training as a random subset of the batch. Consequently, we further constrain samples in a batch to correspond to the same interacting group (see Section~\ref{sec:methodology} for the underlying meta-learning intuition). For the same reason, we also ensure that a batch contains unique observed sequences so that a single observed sequence does not dominate the aggregation of representations over context. This is because a single observed sequence has multiple associated future sequences at different offsets, and could show up multiple times in a batch through random sampling if not handled explicitly.

We optimize the models using Adam \cite{kingmaAdamMethodStochastic2017}. For the NP and SP-MLP models we use a batch size of $128$, an initial learning rate of $3\cdot 10^{-5}$, a weight decay of $5\cdot 10^{-4}$, and a dropout rate of $0.25$. For the MLP-GRU models we use a batch size of $64$, an initial learning rate of $10^{-5}$, and a weight decay of $10^{-3}$. The entire system was implemented using Pytorch \cite{pytorch} and Pytorch Lightning \cite{pytorch-lightning}. Every model was trained on a single NVIDIA GPU on an internal cluster depending on availability; one of Geforce GTX 970 (4~GB) or 1080 (8~GB), or Quadro P4000 (8~GB).

We validate the hyperparameters using $5$-fold cross-validation, in the \textit{random} context regime. At test, we use the same context sequences across models for a fair comparison. All testing was done with a batch size of $128$ for consistency. The errors in mean are computed after destandardizing the location dimensions (orientation is already denoted by a unit quaternion, and therefore not standardized). The predicted std. deviations are scaled by the same value as the predicted means during destandardization.

\section{Distinguishing Forecasting in Focused and Unfocused Interactions: A Meta Discussion} \label{app:review}
Free-standing conversations are an example of what social scientists call \textit{focused interactions}, said to arise when a \say{group of persons gather close together and openly cooperate to sustain a single focus of attention, typically by taking turns at talking} \cite[p.~24]{goffmanBehaviorPublicPlaces1966}.
On the other hand, \textit{unfocused interactions} occur when information is implicitly passed between individuals that happen to be in each other's presence by circumstance, such
as pedestrians walking in proximity.
One practical challenge of forecasting cues in focused interactions stems from the subtlety and sparsity of motion in recorded data. A common assumption is to use head pose as a proxy for gaze \cite{vazquezMaintainingAwarenessFocus2016,rienksSpeakerPredictionBased, farenzenaSocialInteractionsVisual2013, sanghvi2019mgpi, alameda-pinedaAnalyzingFreestandingConversational2015, tanMultimodalJointHead2021}. In real-world data, however, attention shifts through changes in gaze are not always accompanied by similar head rotations \citep[Fig.~5]{baRecognizingVisualFocus2009}. However, gaze is hard to record during group interactions in the wild with reasonable accuracy in a non-invasive manner. Even with the technology to do so (e.g. using onboard sensors on a social robot interaction partner), the question of whether recording faces is privacy-preserving is an ongoing discussion in the community. Moreover, intrusive sensing or non-human partners might also invalidate the naturalness of interaction behaviors (ecological validity). The consequence of not recording gaze is that in dyadic and triadic configurations where people are within each other's field of vision, the recorded movements (only from head and body) are even more subtle since attention shifts are predominantly achieved through gaze changes. This subtlety of motion in recorded data further distinguishes forecasting in conversations from the unfocused setting of pedestrian (or vehicle) trajectories. While some modeling techniques might be computationally applicable in both scenarios, the data stream in pedestrian trajectory settings (locations) can be comparatively more dynamic than the data streams in conversations (e.g. pose). It is important for researchers to be aware of such nuances while interpreting results for downstream applications (see Section~\ref{sec:discussion}).

\end{document}